\def\documenttype{arxiv}

\def\arxiv{arxiv}

\ifx\documenttype\arxiv
  \documentclass{article}
  \usepackage{style/arxiv}
  \usepackage{amsmath}
  \usepackage{natbib}
  \usepackage[usenames, dvipsnames, hyperref]{xcolor}
  \definecolor{MyBlue}{RGB}{0, 20, 115}
  \usepackage[colorlinks=true, citecolor=MyBlue, linkcolor=MyBlue, urlcolor=MyBlue, breaklinks=true]{hyperref}
  \usepackage[capitalize]{cleveref}
  \usepackage[utf8]{inputenc}
  \usepackage[T1]{fontenc}
\fi

\usepackage{lineno}

\usepackage{verbatim}
\usepackage{xspace}
\usepackage{setspace}

\usepackage{caption}
\usepackage{booktabs}
\usepackage{varwidth}
\usepackage{subcaption}
\usepackage{fancyhdr}
\usepackage{multirow}
\usepackage{enumitem}
\setlist{nosep, leftmargin=14pt}
\usepackage{url}      
\usepackage{color,soul}      

\usepackage{graphicx}
\usepackage[
usenames,
dvipsnames,
hyperref]{xcolor}

\usepackage{amsmath}
\usepackage{amssymb}
\usepackage{amsfonts} 
\usepackage{bm}
\usepackage{dsfont}
\usepackage{mathtools}

\usepackage{siunitx}

\usepackage[breaklinks=true]{hyperref}
\usepackage[capitalize]{cleveref}

\crefname{equation}{Eq.}{Eqs.}
\crefname{figure}{Fig.}{Figs.}
\crefname{section}{Sec.}{Secs.}
\crefname{table}{Tab.}{Tabs.}
\crefname{appendix}{Appx.}{Appx.}
\Crefname{equation}{Equation}{Equations}
\Crefname{figure}{Figure}{Figures}
\Crefname{section}{Section}{Sections}
\Crefname{table}{Table}{Tables}
\Crefname{appendix}{Appendix}{Appendices}

\newcommand{\Sim}{\operatorname{S_c}}

\newcommand{\mmu}[1]{\mbox{#1 {\textmu}m}\xspace}
\newcommand{\mcmu}[1]{\mbox{#1 {\textmu}m³}\xspace}

\newcommand{\mpx}[1]{\mbox{#1 pixels}\xspace}

\newcommand{\clip}{\mbox{CL-2D}\xspace}
\newcommand{\clcs}{\mbox{CL-3D}\xspace}
\newcommand{\simclr}{\mbox{Pre-trained (ImageNet)}\xspace}

\newcommand{\mpli}{\mbox{3D-PLI}\xspace}

\DeclareMathOperator*{\img}{\mathrm{i}}

\ifx\documenttype\arxiv
  
  \renewcommand{\cite}[1]{\citep{#1}}
\fi

\title{Self-Supervised Representation Learning for Nerve Fiber Distribution Patterns in 3D-PLI}

\ifx\documenttype\arxiv
  
\author{
  Alexander Oberstrass$^{1,2}$\;
  Sascha E. A. Muenzing$^{1}$\;
  Meiqi Niu$^{1}$\;
  Nicola Palomero-Gallagher$^{1,3}$\;
  \\
  \bf{
  Christian Schiffer$^{1,2}$\;
  Markus Axer$^{1,4}$\;
  Katrin Amunts$^{1,3}$\;
  Timo Dickscheid$^{1,2,5}$\;}
  \\[5mm]
  $^{1}$ Institute of Neuroscience and Medicine (INM-1), Research Centre Jülich, Germany \\
  $^{2}$ Helmholtz AI, Research Centre Jülich, Germany \\
  $^{3}$ Cécile \& Oskar Vogt Institute of Brain Research, Medical Faculty and University Hospital Düsseldorf, \\Heinrich Heine University Düsseldorf, Germany \\
  $^{4}$ Department of Physics, University of Wuppertal, Germany \\
  $^{5}$ Institute of Computer Science, Faculty of Mathematics and Natural Sciences, \\Heinrich Heine University Düsseldorf, Germany
 }

\fi

\begin{document}
  
  \maketitle

  
  \begin{abstract}
A comprehensive understanding of the organizational principles in the human brain requires, among other factors, well-quantifiable descriptors of nerve fiber architecture.
Three-dimensional polarized light imaging (3D-PLI) is a microscopic imaging technique that enables insights into the fine-grained organization of myelinated nerve fibers with high resolution.
Descriptors characterizing the fiber architecture observed in \mpli would enable downstream analysis tasks such as multimodal correlation studies, clustering, and mapping.
However, best practices for observer-independent characterization of fiber architecture in \mpli are not yet available.
To this end, we propose the application of a fully data-driven approach to characterize nerve fiber architecture in \mpli images using self-supervised representation learning.
We introduce a \textit{3D-Context Contrastive Learning } (\clcs) objective that utilizes the spatial neighborhood of texture examples across histological brain sections of a 3D reconstructed volume to sample positive pairs for contrastive learning.
We combine this sampling strategy with specifically designed image augmentations to gain robustness to typical variations in \mpli parameter maps.
The approach is demonstrated for the 3D reconstructed occipital lobe of a vervet monkey brain.
We show that extracted features are highly sensitive to different configurations of nerve fibers, yet robust to variations between consecutive brain sections arising from histological processing.
We demonstrate their practical applicability for retrieving clusters of homogeneous fiber architecture, performing classification with minimal annotations and query-based retrieval of characteristic components of fiber architecture such as U-fibers.

\end{abstract}
\begin{keywords}
{deep learning, contrastive learning, fiber architecture, polarized light imaging, occipital lobe, vervet monkey brain}
\end{keywords}


  \section{Introduction}
\label{sec:intro}

Decoding the human brain requires analyzing its structural and functional organization at different spatial scales, including cytoarchitecture and fiber architecture at microscopic resolutions \cite{amunts2015,axer2022}.
Three-dimensional polarized light imaging (\mpli) \cite{axer2011} is an imaging technique that reveals the fine-grained configuration and 3D orientation of myelinated nerve fibers in both gray and white matter with micrometer resolution.
\mpli thus establishes a link between microscopic myeloarchitecture and dMRI-based structural connectivity at the macro- and mesoscopic scale \cite{zilles2016,caspers2019}.
\mpli images provide detailed visual information for obtaining maps of fiber architecture at different scales.
Based on \mpli images, previous work demonstrated the detection of myelinated pathways and delineation of subfields in the human hippocampus \cite{zeineh2017} as well as the identification of fiber tracts and visual areas in the vervet monkey visual system \cite{takemura2020}.

Polarized light imaging allows processing of whole-brain tissue sections and enables scanning of large tissue stacks \cite{axer2020a,axer2020b,howard2023}.
However, interpretation and analysis of the complex information provided by \mpli requires substantial expertise that cannot scale to the vastly increasing amount of data produced by recent high-throughput devices.
Moreover, \textit{automated} large-scale analysis of fiber architecture at the resolution provided by \mbox{\mpli} is challenging due to the complexity and high dimensionality of the data.
In order to use data analysis algorithms, a suitable lower-dimensional feature representation of \mbox{\mpli} textures is needed.
Ideally, features in this representation are highly expressive for different fiber configurations while being robust against other sources of variation, such as histological processing effects and the relative 3D orientation of image patches.
Such features, however, are difficult to derive, and we hypothesize that an efficient representation cannot be manually engineered.

Over the last years, deep learning methods have become prevalent in analyzing images in related fields such as histopathology \mbox{\cite{dematos2021}}, as they are able to learn representations from pure data.
While annotations of fiber configurations in \mbox{\mpli} are not yet available at the scale required for supervised deep learning, we do have access to large amounts of unlabeled data.
Recent advances in self-supervised representation learning suggest using contrastive learning \cite{hadsell2006,oord2018} to learn distinctive representations from unlabeled training data.
The training objective here is to represent similar instances (positive pairs) as close points in the embedding space while pushing dissimilar instances (negative pairs) apart to prevent representational collapse.
While other methods to prevent representational collapse have been proposed as well, such as clustering \cite{caron2020},
distillation \cite{grill2020,chen2021b},
information maximization \cite{zbontar2021} or
variance preservation \cite{bardes2021}, contrastive learning of visual representations by application or adaptation of SimCLR \cite{chen2020a} and MoCo \cite{he2020} is still popular in medical image analysis \cite{chen2022,krishnan2022}.
Due to its simplicity, we build on the SimCLR framework \cite{chen2020a}.
An application of SimCLR in cytoarchitectonic brain mapping was recently performed for histological images \cite{schiffer2021b}.
A main challenge they discovered was the tendency of models to focus more on anatomical landmarks than on features descriptive of cytoarchitecture when creating positive pairs based on data augmentations of the same image.
To overcome this effect, they employ a supervised contrastive loss \cite{khosla2020} by defining positive pairs based on same labels and sample pairs within each brain area.

Several self-supervised learning methods were proposed to learn image representations based on spatial context, which can be used to create correlated views for contrastive learning \cite{oord2018,chen2020a,vangansbeke2021} or to define pre-text tasks \cite{doersch2015,noroozi2016,pathak2016}.
For microscopic imaging, predicting the geodesic distance between image patches along the brain surface \cite{spitzer2018} or the sequence of multi-resolution histopathology images \cite{srinidhi2022} have been proposed as pre-text tasks.
Other approaches leverage the spatial continuity of images by maximizing mutual information between neighboring patches in histological images \cite{gildenblat2019} or satellite images \cite{ji2019}.
They assume textures in spatial proximity to be similar and therefore aim to contrast them with textures in more distant parts of images.

In the present study, we explore self-supervised contrastive learning for inferring descriptive features of local nerve fiber distribution patterns from raw \mpli measurements.
To generate positive pairs of \mpli texture examples, we assume that fundamental properties of local fiber architecture are typically consistent between nearby image patches.
While this assumption is likely violated at boundaries between distinct structural brain areas, we assume that it holds for the largest share of nearby image patches.   
In contrast to previous work \cite{gildenblat2019,ji2019}, instead of utilizing in-plane similarity of images, we use a 3D reconstructed histological volume to access the spatial coordinates of image patches in 3D.
More precisely, we extract positive pairs of image patches at nearby coordinates across tissue sections.
This sampling strategy is motivated by the idea that positive pairs from different tissue sections show independently measured tissue and thus encourage the learning of features that are robust to random variations in the measurement process not descriptive of fiber architecture.
We denote this 3D-informed self-supervised learning strategy as \textit{3D-Context Contrastive Learning} (\clcs).

To verify the validity of the proposed approach, we compare texture representations by different methods on the 3D reconstruction of the occipital lobe of a vervet monkey brain.
We evaluate features based on their descriptive power for different fiber configurations, as well as their robustness to other sources of variation.
Furthermore, we study the relationship between texture features and morphological measures at the macroscopic scale, using a precise automatic cortex segmentation which we developed specifically for \mpli images based on a \mbox{U-Net} model \cite{ronneberger2015}.
To demonstrate the applicability of the learned \clcs features, we show that the features form clusters that reflect different types of fiber architecture, enable classification tasks with minimal labels and are suitable for query-based retrieval of U-fiber structures.

The main contributions of the present study are the following:
\begin{itemize}[itemsep=1mm,leftmargin=.05\textwidth]
    \item We propose a novel \textit{3D-Context Contrastive Learning} (\clcs) strategy to learn a powerful feature embedding for microscopic resolution image patches from \mpli.
    \item We present specific image augmentations for maximizing invariance of learned features with respect to typical variations in \mpli images, increasing feature quality and robustness.
    \item We show a high sensitivity of the resulting \mpli feature embeddings to fundamental configurations of nerve fibers, such as myelinated radial and tangential fibers within the cortex, fiber bundles, crossings, and fannings, as well as cortical morphology.
    \item Using a dataset from a vervet monkey brain, we demonstrate that the learned features are well suited for exploratory data analysis, specifically for finding clusters of similar fiber architecture and retrieving locations with specific architectural properties based on interactively chosen examples.
\end{itemize}

  \section{Materials and Methods}
\label{sec:methods}

\subsection{\mpli measurements from the occipital lobe of a vervet monkey brain}
\label{sec:occipital_pole}

\begin{figure*}[t]
  \centering
  \includegraphics[width=\textwidth]{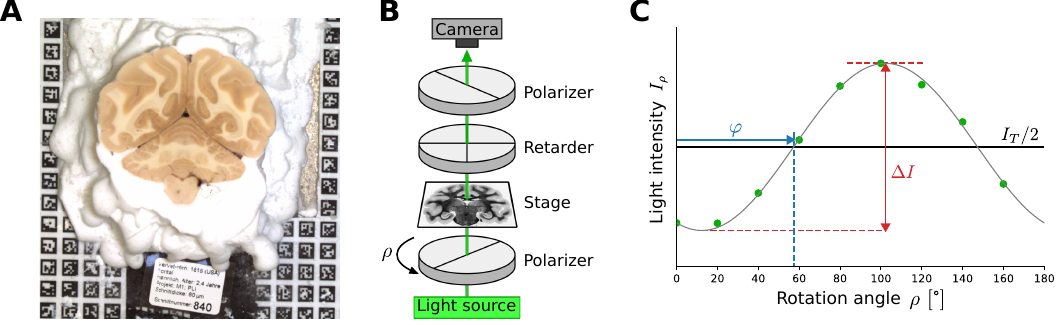}
  \caption{
    Overview of the \mpli data acquisition.
    (A) Blockface images are taken for every section before slicing the mounted tissue block, providing a distortion-free reference for 3D volume reconstruction.
    ARTag markers positioned on the cryotome in the background are used for precise image alignment.
    (B) \mpli measurement setup for the polarizing microscope (LMP-1) consisting of a coherent green light source, a rotating linear polarizer, a specimen stage, a stationary circular analyzer (quarter-wave
    retarder, linear polarizer) and a CCD camera to capture transmitted light intensities.
    (C) Example intensity profile recorded by a single pixel of the CCD camera at 9 polarizer rotation angles $\rho$.
    The profile can be described by a sinusodial curve, parameterized by three modalities: transmittance $I_T$, direction $\varphi$, and retardation $|\sin \delta| = \Delta I / I_T$.
  }
  \label{fig:pli_overview}
\end{figure*}

\paragraph{Tissue samples.}
For this study, we use a 3D reconstruction of 234 coronal sections from the right occipital lobe of a 2.4-year-old adult male vervet monkey brain (ID 1818) measured with \mpli \cite{takemura2020}.
The brain sample was obtained post-mortem after flush with phosphate-buffered saline in accordance with the Wake Forest Institutional Animal Care and Use Committee (IACUC \#A11-219) and conforming the AVMA Guidelines for the Euthanasia of Animals.
It was perfusion fixed with 4\% paraformaldehyde, immersed in 20\% glycerin for cryo-protection, and frozen at -70 C°.
Sectioning of the frozen brain was performed coronally at \mmu{60} thickness using a large-scale cryostat microtome (Poly-cut CM 3500, Leica, Germany).
Before each cutting step, \emph{blockface images} \cite{axer2011} were taken as an undistorted reference for image realignment using a CCD camera (\cref{fig:pli_overview}A).

\paragraph{\mpli acquisition.}
For \mpli measurement \cite{axer2011, axer2011a, axer2022}, brain sections were scanned using a polarizing microscope (LMP-1, Taorad, Germany), which provides a detailed view of nerve fiber architecture at \mmu{1.3} resolution.
In this microscope setup, sections are placed on a stage between a rotating linear polarizer and a stationary circular analyzer consisting of a quarter-wave retarder and a second linear polarizer (\cref{fig:pli_overview}B).
The setup is illuminated by an incoherent white light LED equipped with a band-pass filter of 550 $\pm$ 5 nm half-width.
Variations in transmitted light intensity are captured using a CCD camera for nine equidistant rotation angles $\rho$ of the rotating linear polarizer covering {180\textdegree} of rotation.
The recorded light intensity variations feature sinusoidal profiles at each pixel (\cref{fig:pli_overview}C), which are determined by the spatial orientation of myelinated nerve fibers. 
Using Jones calculus, a physical description for these profiles can be derived as
\begin{equation}
  I_\rho = \frac{I_T}{2} \cdot \left(1 + \sin(2\rho - 2\varphi) \cdot \sin \delta \right).
  \label{eq:3d_pli}
\end{equation}
Harmonic Fourier analysis can be applied to retrieve parameter maps of transmittance $I_T$, retardation $|\sin \delta|$ and direction $\varphi$ from the profiles \cite{axer2011a}.
The phase shift $\delta$ between the ordinary and the extraordinary ray can be further decomposed as
\begin{equation}
  \delta \approx 2 \pi \frac{t \cdot \Delta n}{\lambda} \cos^2\alpha
  \label{eq:phase_retardation}
\end{equation}
with the cumulative thickness of birefringent tissue $t$, birefringence $\Delta n$, the wavelength of the light source $\lambda$, and nerve fiber inclination angle $\alpha$.
While birefringence $\Delta n$ and wavelength $\lambda$ are kept constant for all pixels, the amount of myelinated nerve fibers, reflected by $t$, varies.
To resolve \cref{eq:phase_retardation} for inclination $\alpha$, a transmittance-weighted model \cite{menzel2022} can be used to estimate $t$.
Fiber inclination and direction information can then be jointly visualized in fiber orientation maps in HSV color space \cite{axer2011}, where the hue value corresponds to in-plane fiber direction $\varphi$ while saturation and value reflect the out-of-plane fiber inclination $\alpha$ (both zero for vertical fibers at $\alpha$ = 90°).

\paragraph{3D registration.}

\begin{figure*}[t]
  \centering
  \includegraphics[width=.95\textwidth]{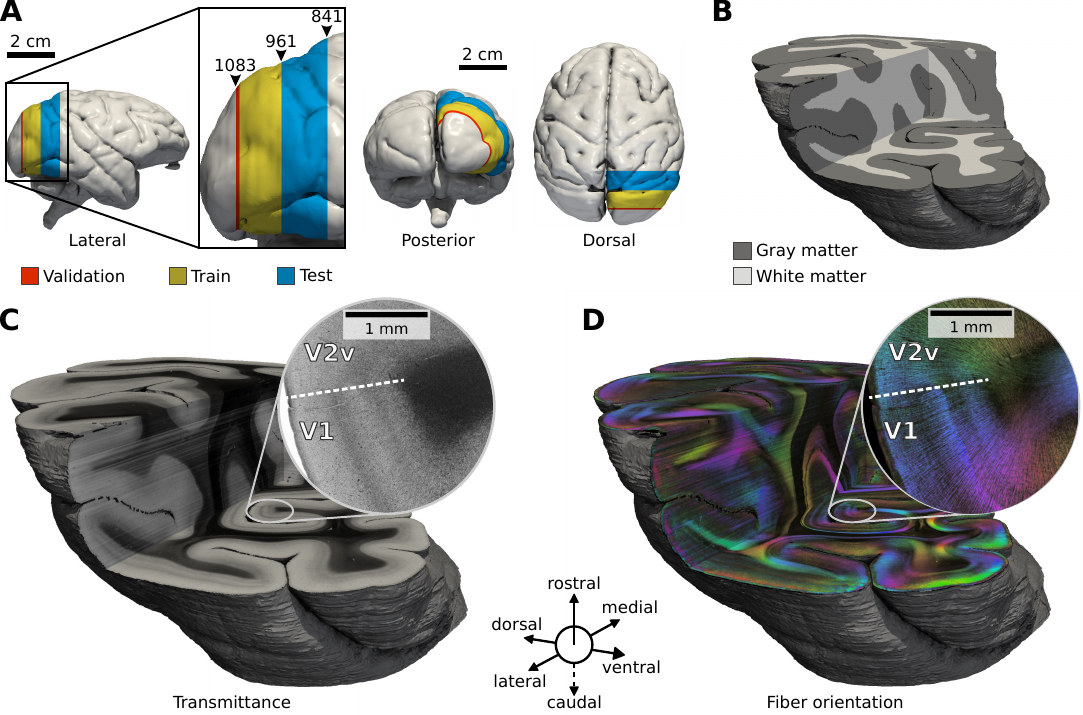}
  \caption{
    3D reconstructed occipital lobe of the right hemisphere of a vervet monkey brain measured with \mpli.
    (A) Localization of the occipital lobe on the surface of the 3D blockface reconstruction.
    Sections used for training (yellow), validation (red), and testing (blue) are color-coded.
    Numbers indicate section numbers.
    (B) 3D volume rendering of segmented cortical gray matter and white matter of the lobe.
    (C) 3D volume rendering for transmittance maps $I_T$ and (D) fiber orientation in HSV color space (hue: direction $\varphi$; saturation, brightness: inclination $\alpha$).
    Zoom-ins highlight the fiber architecture at the border between primary visual cortex (V1) and secondary visual cortex (V2).
    All volumes are masked at the pial boundary shown as a gray surface.
  }
  \label{fig:occipital_volume}
\end{figure*}

To access the three-dimensional context of images, registration of \mpli parameter maps is performed on 234 sections of the right occipital lobe from section 841 to 1083 (\cref{fig:occipital_volume}).
9 sections heavily deformed by histological processing are sorted out and replaced by their nearest neighbors to ensure high-quality 3D reconstruction.
Before registration of \mbox{\mpli} parameter maps, a volume reconstruction of blockface images for the complete brain is performed to serve as an undistorted reference space (\mbox{\cref{fig:occipital_volume}}A).
We use this reference space to correct for distortions from histological processing such as shrinkage or expansion of tissue, and to anchor the occipital lobe in the whole brain context.
For reconstruction of blockface images, ARTag markers are positioned on the cryotome along with the mounted tissue block \mbox{(\cref{fig:pli_overview}A)}.
By identification of the markers using the ARToolKitPlus library \mbox{\cite{wagner2007a}}, blockface images are aligned using affine transformations \mbox{\cite{schober2015}}.
Subsequently, non-linear transformation fields are estimated for alignment of \mpli transmittance maps using the blockface volume as reference, which yields a reconstruction of the overall anatomical shape and topology of the occipital lobe in the \mpli volume space.
The same transformation is used to align all \mbox{\mpli} parameter maps.
However, blockface images do not contain sufficient structural detail for precise alignment of fine structures visible in \mpli such as single fiber bundles and small blood vessels.
Therefore, the blockface alignment is used as initialization for an additional registration step between adjacent \mbox{\mpli} sections to reconstruct coherent 3D fiber tract transitions.
In this step, each section is aligned to its successor and predecessor by symmetric normalization, which combines affine and deformable transformation, maximizing cross-correlation between joint retardation and transmittance images.
The registration is performed iteratively forwards and backwards through the stack of sections, with the first and last section remaining fixed.
All registrations are performed using the \textit{ELASTIX} \cite{klein2010,shamonin2014}, \textit{ANTs} \cite{avants2010,avants2011} and \textit{ITK} \cite{mccormick2014} software packages.
The computed transformation fields are used to warp all \mpli parameter maps into a common volume space (\cref{fig:occipital_volume}C) with a resolution of 31077 $\times$ 28722 $\times$ 243 voxels and a voxel size of \mcmu{1.3 $\times$ 1.3 $\times$ 60}.
In-plane orientation information reflected by direction maps $\varphi$ is preserved by adjusting the 2D rotation components at each pixel estimated by the curl of the transformation field.

\subsection{Data augmentations for \mpli}
\label{sec:transformations}

\begin{figure}[t]
  \centering
  \includegraphics[width=.95\textwidth]{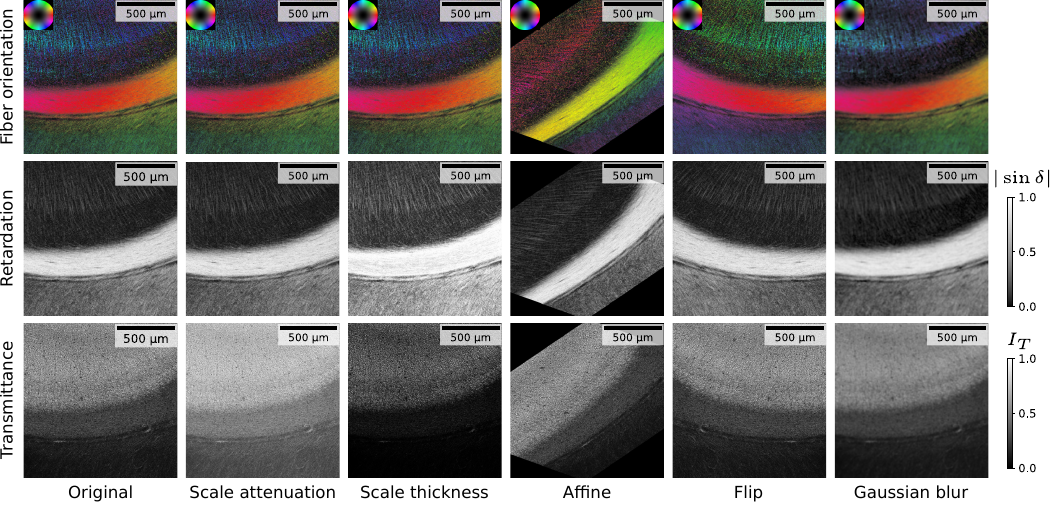}
  \caption{
    Illustration of implemented \mpli data augmentations for an example patch from the calcarine sulcus.
    Images show transmittance $I_T$, retardation $|\sin \delta|$ and fiber orientation in HSV color space (hue: direction $\varphi$; saturation, brightness: inclination $\alpha$).
    The colormap for retardation is scaled with a gamma correction for visibility.
    Parameters for the transmittance weighted model to compute fiber orientations are kept constant for all augmentations.
  }
  \label{fig:data_augmentations}
\end{figure}

Data augmentations are crucial for increasing the diversity of training data.
Self-supervised contrastive learning methods in particular rely on augmentations to learn representations that are more generalizable and robust to variations in the input data \cite{chen2020a}.
It is crucial that augmentation schemes model the expected variability adequately.
In microscopy, for example, similar tissue can exhibit different orientations or intensities between scans.
Since mounted brain sections are not perfectly flat, sharpness variation can occur within a scan, resulting in slightly out-of-focus areas.

In this section, we introduce a set of augmentations specifically designed to reflect typical variations in \mpli images.
We scale attenuation and thickness parameters in the physical model of the measured \mbox{\mpli} signal, perform geometric affine and flip transformations, as well as Gaussian blur.
All augmentations are performed for joint transmittance, direction, and retardation parameter maps.
\mbox{\cref{fig:data_augmentations}} shows example applications of the derived augmentations.

\subsubsection{Modulation of signal parameters}

Parameters of the tissue in the physical model of \mpli might vary (e.g. transparency, thickness) depending on postmortem time, tissue processing, or storage time of the mounted sections.
Here, we provide transformations that can be implemented into \mpli-specific data augmentations that approximate typical variations in the parameters.

\paragraph{Attenuation coefficient.}
The transmitted light intensity of the tissue can vary across image acquisitions due to a change in light attenuation.
Assuming uniform attenuation for simplicity, transmittance $I_T$ can be described by Bouguer-Lambert's law as
\begin{equation}
  \label{eq:att1}
  I_T = I_0 e^{-t\mu}
\end{equation}
for the intensity of incident light $I_0$, section thickness $t$ and attenuation coefficient $\mu$.
Scaling the attenuation coefficient by linear scaling factor $\gamma_a$ as $\mu' = \gamma_a \mu$ results in a scaled transmittance
\begin{equation}
  \label{eq:att2}
  I_T' = I_0 e^{-t\mu'} = I_0 (\frac{I_T}{I_0})^{\gamma_a}~.
\end{equation}
Note that this equation is only an approximation of a real change in $\mu$ due to the simplifying assumption of uniform attenuation.

\paragraph{Section thickness.}
Although the section thickness was held constant throughout all brain sections used in this study, it might vary between data acquisitions of other samples.
To reflect a linear change in thickness parameter $t$ in \mbox{\cref{eq:phase_retardation}}, we scale phase retardation $\delta \propto t$ by a linear scaling factor $\gamma_t$.
For retardation $r = |\sin(\delta)|$ and $\delta' = \gamma_t \delta$ we obtain a scaled retardation
\begin{equation}
  r' = \sin(\gamma_t \arcsin(r))~.
\end{equation}
To adjust the light transmittance $I_T$, we compute scaled $I_T'$ analog to \cref{eq:att2} for scaled thickness $t' = \gamma_t t$ as follows:
\begin{equation}
  \label{eq:thickness}
  I_T' = I_0 e^{-t'\mu} = I_0 (\frac{I_T}{I_0})^{\gamma_t}~.
\end{equation}
Note that this augmentation is also only an approximation to a real change in thickness $t$, as it does not add or remove tissue components from the measurement.

\subsubsection{Resampling}

Many image transformations require resampling of image intensity values.
For \mpli, resampling of the measured intensities from \cref{eq:3d_pli} can be performed as
\begin{equation}
  I'_\rho = \sum_{i} w_i I_{\rho,i}
  \label{eq:resamplesignal}
\end{equation}
by a weighted mean of intensity values $I_{\rho,i}$ with corresponding weights $w_i$, where $\sum_{i} w_i = 1$.
With \mpli parameter maps, however, we work with derivations of the originally measured image intensities and cannot directly resample values for retardation $r = |\sin \delta|$ and direction $\varphi$.
By representing \cref{eq:resamplesignal} as Fourier series and due to the linearity of the Fourier transformation, resampling of the  \mpli parameter maps can be performed through
\begin{align}
  I_T' &= \sum_{i} w_i I_{T,i} \label{eq:resample1} \\
  r' \cdot e^{\img2\varphi'} &= \frac{1}{I_T'}\sum_{i} w_i r_i I_{T,i} \cdot e^{\img2\varphi_i} \label{eq:resample2}~,
\end{align}
by computing $I_T'$ via \cref{eq:resample1} before obtaining $r'$ and $\varphi'$ from \cref{eq:resample2} via decomposition of the right-hand side into magnitude and phase, which correspond to $r'$ and $2\varphi'$, respectively.
The equations are used for all geometric transformations and filters, such as affine transformations or Gaussian blur, that require resampling of \mpli parameter maps.
We use geometric transformations to account for different orientations or distortion of tissue and Gaussian blur to mimic slightly out-of-focus areas.

\subsubsection{Direction correction}

Since \mpli measures the absolute in-plane orientation of nerve fibers, any transformation that changes the geometry of image pixels requires a subsequent correction of direction values.
For applications in diffusion MRI, Preservation of Principal Directions (PPD) \cite{alexander2001,alexander2001a} was introduced to preserve directional information undergoing non-rigid transformations.
While proposed for 3D diffusion tensors, a similar correction mechanism can be introduced for \mpli, where we restrict the transformations to in-plane transformations for simplicity.
We convert direction angles $\varphi$ to cartesian coordinates normalized to one as
\begin{equation}
  \vec{d} =  \begin{pmatrix} \cos{\varphi} \\ \sin{\varphi} \end{pmatrix}
\end{equation}
in order to generate corrected direction angles $\varphi'$ via
\begin{equation}
  \begin{aligned}
    \vec{d}' &= J_f\vec{d} \\
    \varphi' &= \text{atan2}(d_2', d_1')
    \label{eq:fix_dir_resample}
  \end{aligned}
\end{equation}
for non-linear image transformation function $f\colon {\mathbb{R}^2}  \to {\mathbb{R}^2}$, which maps pixel coordinates in the source domain to coordinates in the target domain, and Jacobi matrix $J_f$ of function $f$.
The correction is applied before application of function $f$ to transform the image.
For specific transformations, \cref{eq:fix_dir_resample} can be simplified to more convenient forms.

\paragraph{Rotation.}
For an example of counter-clockwise rotation by arbitrary angle $\theta$, \cref{eq:fix_dir_resample} can be simplified to
\begin{equation}
  \label{eq:fixrot}
  \varphi' = \varphi + \theta~.
\end{equation}

\paragraph{Affine Transform.}
For pixel coordinates $\vec{p} = [x, y]^T$ and an affine transformation composed of translation vector $\vec{t}$ and matrix $A$, the transformation function is given as
\begin{equation}
  f(x,y) = A \vec{p} + \vec{t}~.
\end{equation} 
If inserted into \cref{eq:fix_dir_resample}, a simplified correction mechanism for the affine transformation can be derived as
\begin{equation}
  \begin{aligned}
    \vec{d}' &= A\vec{d} \\
    \varphi' &= \text{atan2}(d_2', d_1')~.
  \end{aligned}
\end{equation} 

\subsection{Cortex segmentation}
\label{sec:cortexsegmentation}

To access brain morphology and distinguish gray and white matter locations, a U-Net model \cite{ronneberger2015} is trained for segmentation of pixels into background (BG), gray matter (GM), and white matter (WM) classes and applied to every \mpli section of the lobe.
For training the model, we create a dataset representing a large variety of textures in \mpli images with minimal labeling effort by employing an active learning strategy in the annotation process.
Rather than annotating complete sections, we manually select 58 square regions of interest (ROIs) of size \mpx{2048} (2.66 mm) from several sections, including sections outside the occipital lobe for a higher variety of examples.
We train a U-Net model using these ROIs as inputs and apply the model to all available sections.
We subsequently select new ROIs based on the most severe misclassifications in the model outputs.
This process is repeated to obtain a growing dataset of 58, 119, 183, 301 and finally 369 ROIs of highly diverse patches capturing different textures across the entire brain.

It should be noted that large parts of the cortex can be segmented at acceptable quality using simple thresholding of transmittance and retardation values \cite{menzel2022}.
Challenging parts, such as an oblique cut border between gray and white matter or an intersecting pial surface within narrow sulci, form only a small fraction of the data, but have a significant impact on matching inner and outer cortical boundaries.
We therefore apply a multi-class implementation of focal loss \cite{lin2017} for the training objective to increase emphasis for the model on challenging examples.
We use \mpli-specific augmentations, as described in \cref{sec:transformations}, for training the cortex segmentation model.

Segmentations of the final model are corrected manually by removing small tissue fragments, extrapolating broken tissue, and filling holes to obtain a topologically correct cortex segmentation.
As a last step, the resulting segmentations for individual \mpli sections are stacked to form a segmented volume of the entire cortex in 3D (\cref{fig:occipital_volume}B).

\subsection{3D context contrastive learning}
\label{sec:contrastive_learning}
\begin{figure*}[t]
  \centering
  \includegraphics[width=.95\textwidth]{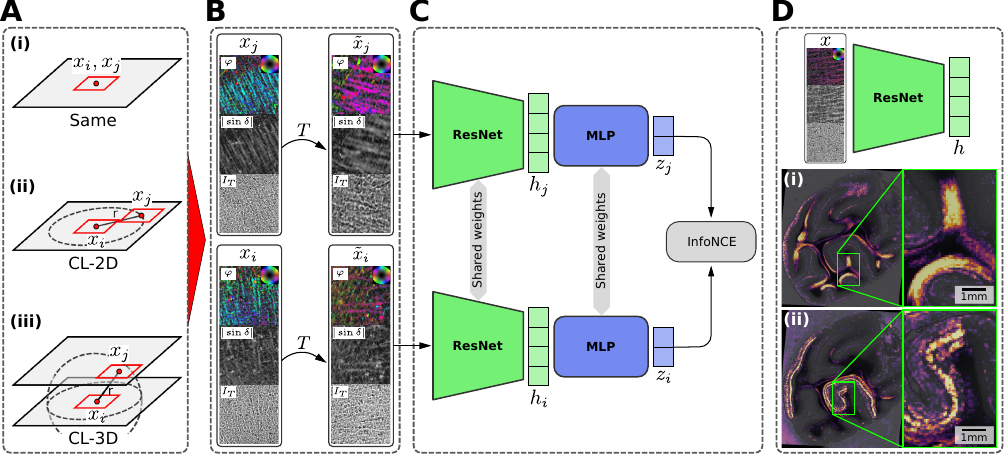}
  \caption{
    Illustration of the proposed 3D context contrastive learning scheme.
    (A) Context sampling performed to obtain correlated views of similar nerve fiber architecture $(x_i, x_j)$ as (i) identical patches (Same), (ii) in-plane shifted patches on a circle with radius $r$ (\clip), or (iii) patches on a sphere with radius $r$ across sections (\clcs).
    (B) Data augmentations $T$ for \mpli are randomly applied to sampled patches to promote learning representations that are robust to typical variations in \mpli measurements.
    Patches are visualized as transmittance $I_T$, retardation $|\sin \delta|$ and fiber orientation map (FOM) in HSV color space (hue: direction $\varphi$; saturation, brightness: inclination $\alpha$).
    (C) SimCLR contrastive learning framework \cite{chen2020a} consisting of a ResNet encoder, hidden features $h_i$ and $h_j$, a fully connected MLP projection head, projections $z_i$ and $z_j$ and InfoNCE loss.
    (D) For inference, the trained encoder is applied on un-augmented patches $x$ to extract \mpli texture features $h$. Whole sections are converted to feature maps using a sliding window approach. Two example feature maps are shown on top of transmittance maps for reference highlighting (i) U-fibers and (ii) primary visual cortex (V1).
  }
  \label{fig:method_overview}
\end{figure*}

Contrastive learning aims to learn robust and descriptive representations of data samples by contrasting similar and dissimilar pairs.
The goal is to learn an encoding function $f$ that groups similar samples closely together in representation space while pushing dissimilar samples apart from each other.
Assuming a reasonable measure of similarity that can be efficiently derived from the data, similar samples are generated as \textit{positive pairs} consisting of an \textit{anchor sample} and a \textit{positive sample} with high similarity, while for \textit{negative pairs} the anchor is combined with a dissimilar \textit{negative sample}.

In this study, we derive similarity from the spatial neighborhood of image patches in a \textit{3D-Context Contrastive Learning} objective.
Given a random location $p^a$ for the anchor sample, we obtain a positive sample from neighborhood location $p^+ = p^a + \Delta p$, where $\Delta p$ is chosen based on two variants of spatial \textit{context sampling} (\cref{fig:method_overview}A):
\begin{enumerate}[label=(\roman*),itemsep=1mm,leftmargin=.05\textwidth]
  \item \textbf{\clip}, where $\Delta p$ is sampled on an in-plane circle with radius $r$ from the same tissue section.
  Setting \mbox{$r$ = 0} is considered a special case, where the anchor sample is also used as positive sample and no context sampling is performed (\textbf{Same}).
  \item \textbf{\clcs}, where $\Delta p$ is sampled on a sphere with radius $r$ and $p^+$ is taken across sections by rounding the sampled coordinate to the nearest available section, but excluding the section from which the anchor sample was taken (i.e. $p^a$ and $p^+$ are always located on different sections).
  Setting \mbox{$r$ = 0} is considered a special case, where $p^+$ refers to the nearest neighbor (\textbf{NN}) at the same in-plane coordinates as $p^a$, but in a random adjacent section.
\end{enumerate}
For $p^a$, only locations with visible tissue are considered to avoid sampling positive pairs containing background only, using the segmentation masks obtained in \cref{sec:cortexsegmentation}.
As \mbox{\clcs} requires access to the spatial relationship of sampling locations between sections, we perform context sampling in the undistorted blockface 3D reference space.
We utilize estimated transformation fields from performing the 3D registration in \mbox{\cref{sec:occipital_pole}} to warp locations in the blockface volume to individual \mbox{\mpli} sections.
Using patches from original \mbox{\mpli} parameters maps instead of registered ones directly has the advantage of including additional variation between positive samples for the contrastive learning objective, such as different orientations of texture.
In addition to the spatial sampling, we perform random augmentations for all samples as detailed in Sec.~\ref{sec:transformations} (\cref{fig:method_overview}B).

For training encoder $f$, we build on the SimCLR contrastive learning framework \cite{chen2020a} (\cref{fig:method_overview}C).
In this specific framework, $N$ augmented positive pairs $(\tilde{x}_i, \tilde{x}_j)$ are randomly sampled for each training step and stored in minibatches of $2N$ total examples $\{\tilde{x}_k\}$.
We refer to the set containing indices $(i,j)$ for all $N$ positive pairs in the minibatch as $\Omega$.
For each positive pair, all $2(N-1)$ other random samples $\{\tilde{x}_k\}_{k\neq i,j}$ are considered negative samples, which originate from random sections and locations.
When the training volume is large relative to sampling radius $r$ of positive pairs, it is unlikely that a negative sample lies in the same spatial neighborhood as the positive one.
Encoder $f$ typically refers to a deep learning model, which yields network activation vectors $h_k = f(\tilde{x}_k)$ as representations.
An additional projection head $g$ is introduced to map the activations $h_k$ to a lower dimensional space of projections $z_k = g(h_k)$, on which contrastive loss is applied.
The training objective is given in terms of the InfoNCE loss \cite{oord2018}:
\begin{equation}
  \ell_{i,j} = - \log \dfrac{\exp({\Sim(z_i, z_j) / \tau})}{\sum_{k=1}^{2N} \mathds{1}_{[k\neq i]} \exp({\Sim(z_i, z_k) / \tau})}~,
  \label{eq:infonce}
\end{equation}
where $z_k = g(f(\tilde{x}_k))$ and similarity metric $\Sim$ chosen as the cosine similarity with temperature parameter $\tau$.
The former per-sample loss is accumulated into a total loss as
\begin{equation}
  \mathcal{L} = \frac{1}{2N} \sum_{(i,j) \in \Omega} \ell_{i,j} + \ell_{j,i}~.
  \label{eq:loss}
\end{equation}
After training, the projection head is discarded and for inference only representations $h=f(x)$ on un-augmented samples $x$ are used (\cref{fig:method_overview}D).

\subsection{Model training}
\label{sec:modeltraining}

\paragraph{Model Architecture.}
For encoder $f$, we use a ResNet-50 \cite{he2016} model with 3 input channels, removing the last fully-connected layer.
The original ResNet-50 encoder outputs 2048 feature channels, which is well evaluated on natural images \mbox{\cite{deng2009}}.
However, training the model at full capacity on our data systematically resulted in high activations to infrequent but highly pronounced structures such as tangential cut radial fibers.
Such structures made it trivial to solve the contrastive learning objectivate based on spatial similarity.
Therefore, we reduce the number of features for all blocks in the ResNet-50 architecture to 1/8 to limit the encoder capacity, which results in 256-dimensional hidden representations, preventing the model from overfitting to these specific structures.
We choose this dimensionality as a trade-off between preventing overfitting and maintaining reasonable model capacity, as bigger models can learn more general features \mbox{\cite{chen2021b}}.
For projection head $g$, we use a two-layer MLP with ReLU activations, hidden feature size of 90 and outputs $z$ of size 32.
To feed \mpli images to the ResNet, parameter maps transmittance $I_T$, direction $\varphi$ and retardation $|\sin \delta|$ are stacked as $x$ = ($I_T$, $\sin \delta \cdot \cos(2 \varphi)$, $\sin \delta \cdot \sin(2 \varphi)$) to the channel dimension, which resolves the cyclic nature of direction values $\varphi$.
We standardize the input channels by running mean and standard deviation over the first 1\,024 batches during training.

\paragraph{Implementation.}
We use \textit{PyTorch} \cite{NEURIPS2019_9015}, \textit{PyTorch Lightning} \cite{borovec2022}, and \textit{Hydra} \cite{yadan2019} frameworks using the \textit{Quicksetup-ai} \cite{mekki2022} template for building our model.
Data augmentations (cf.~\cref{sec:transformations}) are implemented using the \textit{Albumentations} \cite{info11020125} framework.
Training is conducted using a distributed data-parallel strategy on 4 Nvidia A100 GPUs with synchronized batch normalization statistics \cite{ioffe2015} on the supercomputer JURECA-DC at the Jülich Supercomputing Centre (JSC) \cite{thornig2021}.

\paragraph{Data sampling.}
We sample square patches of size \mpx{192} (\mmu{253}) as anchor samples from random locations within the training volume, excluding background using the previously generated cortex segmentation.
For each anchor sample, we take positive samples from a random location in spatial proximity, depending on the chosen definition of spatial similarity.
As we sample patch locations on the fly, we do not have a fixed dataset size but define an \textit{epoch} as the sampling of 512 $\times$ 512 = 262\,144 positive pairs.
Per training step, we take 512 anchor samples and positive samples and process them evenly split on the 4 GPUs.

\paragraph{Data augmentation.}
For all samples, we apply an affine transformation (scaling from [0.9, 1.3] on each axis, rotation from [-180°, 180°], and shearing from [-20°, 20°] on each axis) with linear interpolation and subsequent center cropping to crops of size \mpx{128} (\mmu{169}) to eliminate padding effects.
Subsequently, we perform random flipping on center crops and scale relative thickness $t$ (\cref{eq:att2}) and the attenuation coefficient $\mu$ (\cref{eq:att1}) each by random scaling from a logarithmic distribution with basis 2 from [-1, 1].
As a last augmentation, we perform Gaussian blur with a probability of 50\% and $\sigma$ from [0.0, 2.0].

\paragraph{Training.}
All augmented crops are fed to the encoder model and projection head in order to minimize the loss in \cref{eq:loss}.
We use Adam optimizer \cite{kingma2017} with a learning rate of 10\textsuperscript{-3}, a weight decay of 10\textsuperscript{-6} and default parameters \mbox{$\beta_1$ = 0.9}, \mbox{$\beta_2$ = 0}.999 and \mbox{$\epsilon$ = 10\textsuperscript{-8}}.
For the choice of temperature parameter $\tau$ in \cref{eq:infonce}, we follow the optimal choice of \mbox{$\tau$ = 0.5} reported by \cite{chen2020a} when training until convergence.
We apply the same loss for training and validation.
All models are trained until convergence if the validation loss does not reduce for more than 50 epochs, which takes between 195 (2D context) and 400 (3D context) epochs.

\paragraph{Inference.}
After training, model weights are frozen and inference is performed on complete sections using the trained ResNet encoder, discarding the projection head.
Each section is converted into feature maps using a sliding window approach by dividing \mpli parameter maps into tiles of size \mpx{128} (\mmu{169}) with 50\% overlap.
The overlap is chosen to better represent pixels at the edges of patches that would otherwise lie on the boundary between adjacent patches.
We extract a 256-dimensional feature vector for each tile without applying the data augmentation used in training.
Extracted feature vectors are reassembled into feature maps with reduced in-plane resolution of \mmu{84.4} per pixel compared to original input parameter maps, but 256 feature channels characterizing the local texture content.
Compared to the section thickness of \mmu{60}, this makes the feature voxels approximately isotropic.

\subsection{Classical texture features}
\label{sec:baseline_glcm}

The present approach enables learning of texture features specifically for \mpli parameter maps.
As baselines for texture analysis, we apply classical first-order histogram features (mean, variance, skewness, kurtosis, entropy), Grey-Level Co-occurrence Matrices (GLCM) \mbox{\cite{haralick1973}}, Local Binary Patterns (LBP) \mbox{\cite{ojala2002}} and a combined set of all of their features.
These are well-established approaches to represent textures in medical imaging such as CT, MR, PET \mbox{\cite{scalco2017}} and histopathology \mbox{\cite{dematos2021}}. 
We compute texture features for whole sections using the same sliding window approach introduced in Sec.~\ref{sec:modeltraining} by dividing \mpli parameter maps into tiles of size \mpx{128} (\mmu{169}) with 50\% overlap.

Direction maps $\varphi$ represent the absolute orientation of fibers within the imaging plane.
Since we aim to find texture representations that are independent of their absolute orientation, we are more interested in local patterns of $\varphi$ than in their absolute values.
We use the Sobel operator as a first derivative filter to highlight image edges and eliminate absolute values of $\varphi$.
To filter direction values $\varphi$, we need to resolve their circular nature.
Therefore, we represent direction angles $\varphi$ in polar form as complex numbers
\begin{equation}
  z = \cos(2\varphi) + i \sin(2\varphi)
\end{equation}
and apply Sobel filtering as
\begin{equation}
  G_x =  K_x * z \hspace{3mm} \text{and} \hspace{3mm} G_y =  K_y * z
\end{equation}
with convolution operator $*$ and Sobel filter kernels $K_x$ and $K_y$.
We aggregate the filtered images as
\begin{equation}
  \hat{\varphi} = \vert G_x + G_y \vert / 12~,
\end{equation}
where $\vert . \vert$ extracts the magnitude of complex numbers, and dividing by 12 normalizes the filtered values to [0, 1].

For histogram features, we compute normalized histograms with 128 bins for each of the parameter maps $I_T$, $\sin \delta$ and $\hat{\varphi}$.
From the histograms, we compute mean, variance, skewness, kurtosis and entropy as features.
Features for all parameter maps are concatenated, resulting in a total of 15 histogram features.

To extract LBP features, we compute local binary patterns for each patch by dividing the angular space into 8 points with multiple radii [1, 2, 3] to define the local neighborhood of texture.
We compute normalized histograms with 10 bins of LBP values for each radius and parameter map and concatenate them into a feature vector with 90 features.

We compute normalized and symmetric GLCMs for 32 equally spaced bins of parameter maps for distances [1, 2, 4] and angles [0, $\pi$/4, $\pi$/2, 3$\pi$/4].
From each GLCM, we compute contrast, correlation, energy, and homogeneity as features \cite{haralick1973}.
We concatenate the features for all parameter maps and distances while averaging over the angles to make the features robust to rotations, resulting in 36 total features.

In addition, we include a comprehensive combined set of all classical texture features (Histogram, LBP, GLCM) as a baseline.

\subsection{Pre-trained encoder on ImageNet}
\label{sec:baseline_pretrained}
In recent years, there has been increasing interest in using pre-trained deep learning models as feature extractors in histopathology \mbox{\cite{dematos2021}}, a domain very close to ours.
Here, we observe that many recent studies analyzing histopathological images use encoders pre-trained on ImageNet \mbox{\cite{deng2009}} for feature extraction and downstream analysis \cite{breen2024,wu2024,liu2024}.
Therefore, we complement the classical baselines with a pre-trained ResNet-50 \mbox{\cite{he2016}} encoder, which has been trained on images from the ImageNet dataset using the SimCLR contrastive learning objective \mbox{\cite{chen2020a}}.

As the ResNet-50 model was trained on natural RGB images, we use two types of images generated from \mbox{\mpli} parameter maps to visualize the fiber architecture:
1. Transmittance maps, stacked in the color channel dimension, to create grayscale RGB images, and
2. fiber orientation maps (FOM), which can be directly fed to the model.
In contrast to the width-reduced ResNet-50 architecture we use for \mbox{\clcs} and \mbox{\clip}, this model has full capacity and produces 2048 features for image patches of 128 pixels (\mbox{\mmu{169}}) size.
For the creation of feature maps, we use the same sliding window approach as in the inference of \mbox{\clip} and \mbox{\clcs} (\mbox{\cref{sec:modeltraining}}).

  \section{Experiments and Results}
\label{sec:results}

We train \clip and \clcs models using the data from the occipital lobe of a vervet monkey brain described in Sec.~\ref{sec:occipital_pole}. 
We split the volume into sections for training (\#962 - \#1077), sections for validation (\#1078 - \#1083), and sections for testing (\#851 - \#961) as shown in \cref{fig:occipital_volume}A.
We evaluate the feature representations produced by the models regarding their descriptive power for different fiber configurations, spatial consistency, and applicability for downstream tasks.
In particular, we show that classification of texture into tissue classes requires less annotations based on \mbox{\clcs} and \mbox{\clip} features than other methods.
We compare the extent to which features of different approaches can be related to brain morphology and their robustness to variations between sections.
We investigate the main factors of variation specifically for \clcs features and demonstrate that they lend themselves to interactive data exploration and identification of nerve fiber architecture in large volumes of \mpli data.
All experiments are evaluated exclusively on features extracted from sections not included in training.

\subsection{Linear evaluation of features with minimal labels}

A common approach to assess the quality and robustness of feature representations is to perform linear evaluation \mbox{\cite{oord2018,chen2020a}} on a given classification task.
For the linear evaluation protocol, a simple linear classifier is trained on top of features extracted for each data sample.
Being able to perform the classification task with a simple linear model indicates a good discrimination of classes in feature space and thus high-quality features.
In addition, we analyze the robustness of features in a weakly-supervised setting by providing the classifier with increasing amounts of labeled training examples, starting with only a few per class.
These examples scale with human annotation effort involved in creating the dataset.
Self-supervised \mbox{\clcs} and \mbox{\clip} encoders are still trained on the full amount of unlabeled training data, which does not require human annotations and is therefore available in a much higher quantity.

For the classification task, we use the training data acquired for the cortex segmentation performed in \mbox{\cref{sec:cortexsegmentation}}, which segments \mbox{\mpli} images into three classes: background (BG), gray matter (GM), and white matter (WM).
We divide the annotated ROIs into 228 ROIs (from 42 sections located caudal to the central sulcus) for training and 141 ROIs (from 19 sections encompassing the prefrontal cortex) for testing.
This train/test split ensures that section IDs for testing the classifier do not contain sections from the occipital pole used for training the self-supervised models and allows testing the generalizability of features across the brain.
To perform classification on individual texture patches, we extract square patches of 128 pixels (\mbox{\mmu{169}}) on a regular grid per ROI and assign each patch the label of the most frequent class in the segmentation mask.
The resulting dataset comprises many different textures with slightly unbalanced class distributions for both training (10696 WM, 32613 GM, and 12829 BG patches) and testing (7065 WM, 19577 GM, and 7589 BG patches).

For the linear classifier, we use a logistic regression classifier in a one-versus-rest scheme.
We compare the classification performance by computing macro F1 scores across classes, and calculate the significance of these scores by computing standard error over 50 independent fits of the classifier on random subsets of training examples.

\subsubsection{Evaluation of features by different methods}

\begin{figure}[t]
  \centering 
  \includegraphics[width=.78\textwidth]{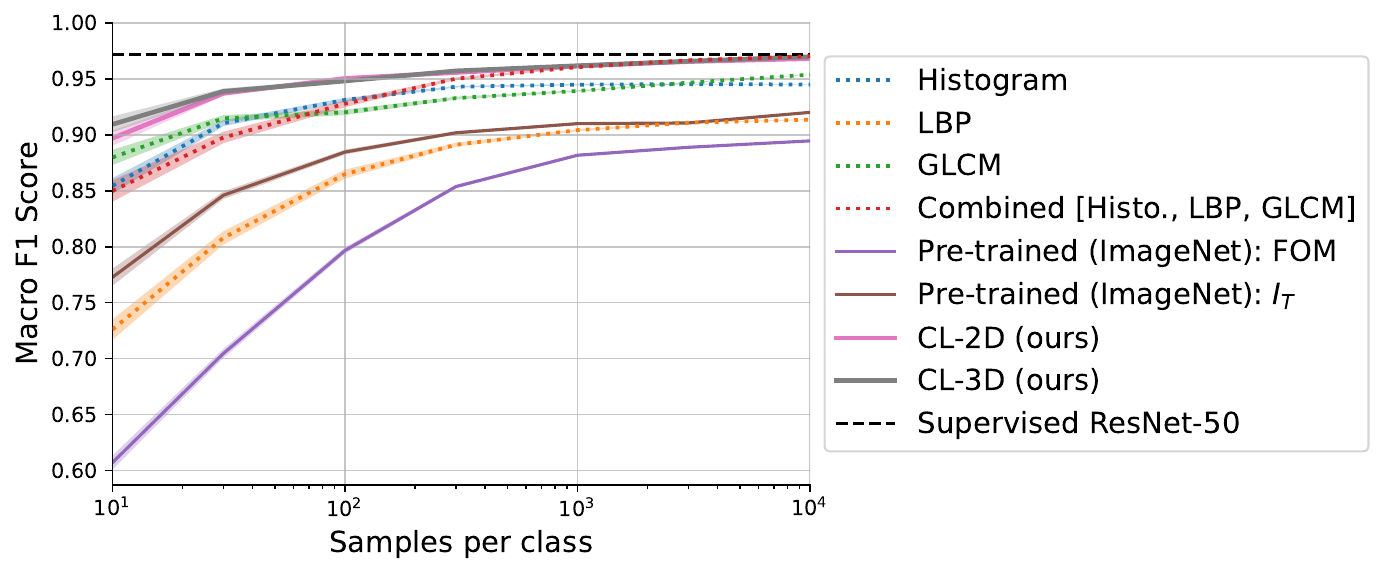}
  \caption{
    Comparison of different feature extraction methods under the linear evaluation protocol.
    A simple linear classifier is fitted on extracted features with an increasing number of labeled samples per class to classify texture patches as gray matter, white matter or background.
    With minimal samples per class provided, \mbox{\clcs} and \mbox{\clip} features perform best, demonstrating highest robustness across the brain.
    Using the full number of samples per class, \mbox{\clcs}, \mbox{\clip} and a combination of classical texture features (Combined) all match the performance of a ResNet-50 model trained specifically on this task, indicating a high-quality feature space of these methods.
    Shaded areas mark standard error over 50 independent fits of the classifier on randomly selected samples.
  }
  \label{fig:lin_ev_methods}
\end{figure}

We compare \mbox{\clcs} and \mbox{\clip} features with classical texture features and a pre-trained ResNet-50 encoder on ImageNet under the linear evaluation protocol.
Additionally, we report the performance of a supervised ResNet-50 classifier as a reference, specifically trained on the classification task using the full training dataset.
For this model, we used a class-weighted cross-entropy loss with Adam optimizer in the same setting as described in \mbox{\cref{sec:modeltraining}}.
The model was trained for 416 epochs until convergence of validation accuracy, using a random 80/20 train/validation split.

With minimal training examples, results in \mbox{\cref{fig:lin_ev_methods}} show a clear lead in classification performance by our \mbox{\clip} and \mbox{\clcs} models.
Both models achieve macro F1 scores of 0.94 with only 30 random samples per class to fit the linear classifier.
By using 10k samples per class, \mbox{\clcs}, \mbox{\clip} and a combination of all classical texture features (Combined) match the F1 score of the supervised ResNet-50 model (0.97).
All other methods, including each individual classical texture descriptor, show lower F1 scores throughout all numbers of samples per class.

\subsubsection{Effect of data augmentations for \mbox{\mpli} on texture features by \mbox{\clcs} and \mbox{\clip}}
\label{sec:lin_ev_ablation}

\begin{figure}
  \centering
  
  \begin{subfigure}{0.48\textwidth}
      \centering
      \includegraphics[width=\textwidth]{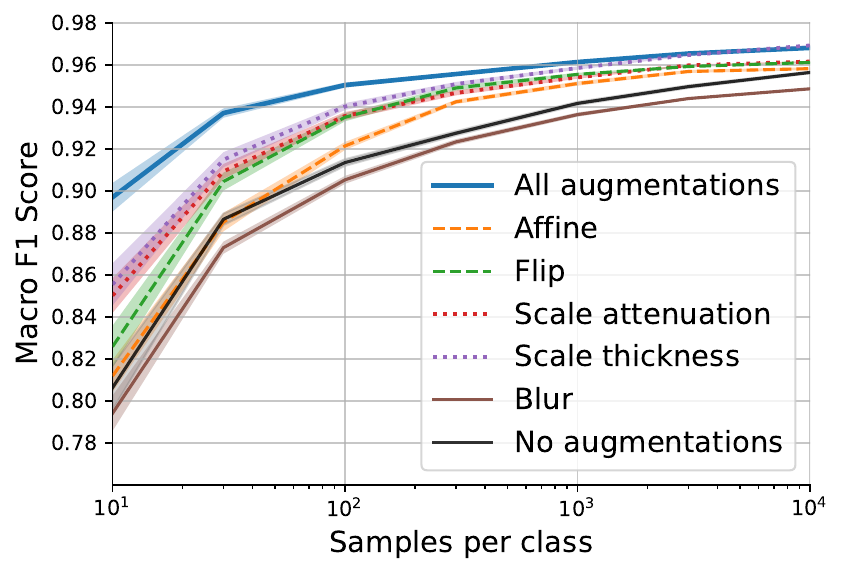}
      \caption{CL-2D}
      \label{fig:lin_ev_cl2d}
  \end{subfigure}
  \hfill
  \begin{subfigure}{0.48\textwidth}
      \centering
      \includegraphics[width=\textwidth]{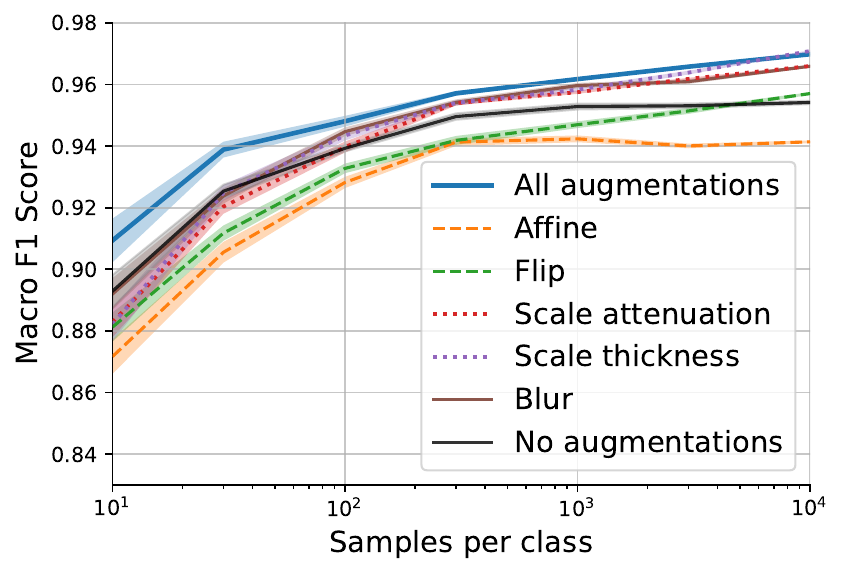}
      \caption{CL-3D}
      \label{fig:lin_ev_cl3d}
  \end{subfigure}
  \caption{
    Impact of different classes of data augmentations on quality and robustness of extracted features by the proposed models (a) \mbox{\clip} and (b) \mbox{\clcs}.
    Quality and robustness are evaluated using the linear evaluation protocol under an increasing number of labeled samples per class.
    A simple linear classifier is fitted on extracted features to differentiate texture patches as gray matter, white matter, or background.
    Macro F1 scores are presented for each model across different augmentation sets.
    Models trained with all augmentations achieve the highest robustness and quality of features, surpassing those trained with individual augmentations or without any augmentations. 
    Shaded areas indicate the standard error over 50 independent fits of the classifier on random training samples.
  }
  \label{fig:lin_ev_cl}
\end{figure}

In \mbox{\cref{sec:transformations}}, we introduce data augmentations specifically designed for \mbox{\mpli} parameter maps and use them in the training of \mbox{\clcs} and \mbox{\clip} models.
To test the effect of these augmentations on feature quality and robustness, we perform linear evaluation of \mbox{\clcs} and \mbox{\clip} models trained without any augmentation, with every single augmentation, and with all augmentations combined.

As shown in \mbox{\cref{fig:lin_ev_cl2d}}, a \mbox{\clip} model trained without data augmentations clearly underperforms compared to models trained with any individual data augmentation, except the blur augmentation, which appears to degrade feature representations for this classification task.
Among models trained with individual augmentations, color distortions (modulation of section thickness and the attenuation coefficient) yield the best results.
A combination of all augmentations together performs the best overall.

Results in \mbox{\cref{fig:lin_ev_cl3d}} demonstrate that \mbox{\clcs} benefits most from color distortions.
Models trained with geometric transformations only, such as affine and flip augmentations, perform worse than a model trained without augmentations.
Excluding them from the full set of augmentations, however, does not improve performance (see \mbox{\cref{fig:lin_ev_geom_cl3d}} in appendix).
Using all introduced augmentations during training leads to the best results for \mbox{\clcs}.

\subsection{Main factors of variation in the learned representations}
\label{sec:pca}
\begin{figure*}[p]
  \centering
  \includegraphics[width=.95\textwidth]{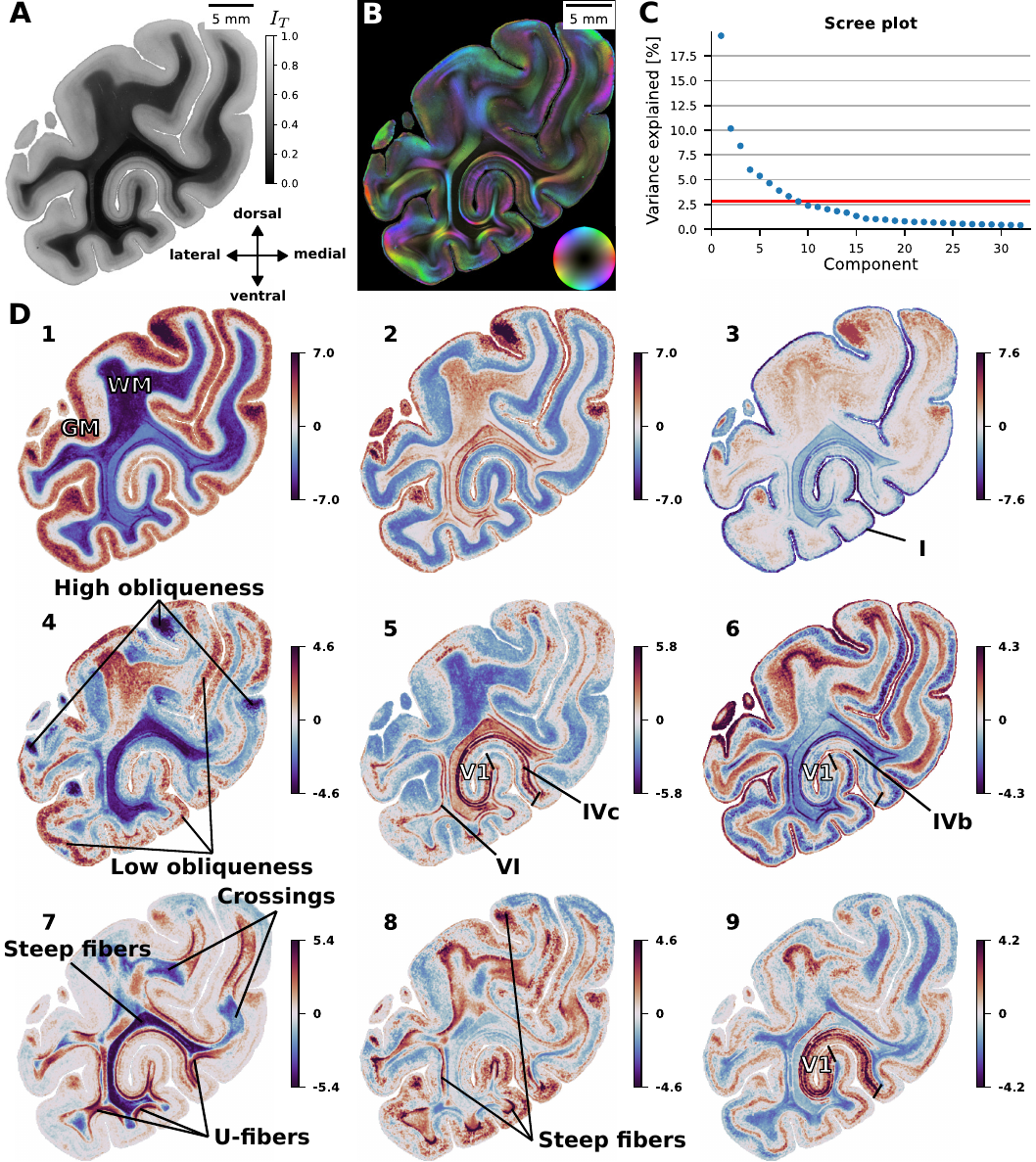}
  \caption{
    Projection of \mbox{\clcs} texture representations from section 898 onto the 9 PCA components with largest explained variance.
    (A) Transmittance and (B) fiber orientation map (FOM) for the section.
    (C) Scree plot showing the variance explained by the first 32 components. The horizontal red line indicates the variance explained by the 9th component of 2.8\%.
    (D) Color-coded parameter maps of the selected PCA components, with background pixels masked as zero. The maps reveal anatomically plausible structures. 
    GM: gray matter, WM: white matter, V1: primary visual cortex.
  }
  \label{fig:pca}
\end{figure*}

To gain insights into the main factors of variation captured by \clcs features, we perform principal component analysis (PCA) on a random subset of 1 million voxels from the feature maps.
We use the estimated principal axes to project feature channels for the entire dataset onto 9 components with largest explained variance (64.2\% cumulative explained variance), with at least 2.8\% of variance explained per component.
Explained variance refers to the amount of information in the dataset that is retained by each PCA component.

\cref{fig:pca}D (see also \cref{fig:pca_appendix} in the appendix) shows images for the first 9 principal components, which in general terms reveal anatomically plausible structures.
Component (1) shows a clear separation of white matter (WM) and gray matter (GM).
Within GM, higher values indicate layers with a lower density of radial fibers.
Within GM, values in component (2) distinguish between layers with high density of radial fibers (low values) and high density of tangential fibers (high values in the superficial layers, which contain mainly tangential fibers, and values around 0 in the Gennari stripe, which presents both radial and tangential fibers), while high values within WM highlight edges between fiber bundles. 
Low values in component (3) highlight layer I, which contains a high density of tangential fibers, as well as WM structures with high fiber density that run in-plane, such as the Tapetum.
High values indicate tangentially sectioned radial fibers.
Component (4) has low values towards out-of-plane fibers and highly oblique cortex and high values towards superficial GM layers with a low density of radial fibers and low obliqueness. 
Component (5) shows high values for layer IVc in the primary visual area (V1) and layer VI throughout the whole cortex.
In component (6), high values show layer I throughout the cortex as well as layer IVb within V1 (Stria of Gennari), i.e., they highlight GM layers with a high density of tangential fibers.
High and low values in (7) mainly represent WM, with high values indicating U-fibers and other in-plane fibers and low values indicating steep fibers or crossings.
The stratum sagittale (SS) has the lowest values, as here fibers emerge vertically from the plane.
High values in GM mainly highlight tangentially sectioned radial fibers.
High values in (8) also highlight tangentially sectioned radial fibers in GM.
Additionally, they indicate abrupt twisting of flat, in-plane fibers that twist out of plane at the GM/WM transition and show parts of U-fibers that are cut through the plane.
In component (9), layers IVa, IVc and VI within V1 are characterized by high values.

\subsection{Encoding of brain morphology in texture features}
\label{sec:linear_relationships}

\begin{table*}[t]
  \centering
  \caption{
    Proportion of variance in morphological measures cortical depth, white matter (WM) depth, curvature, and obliqueness that can be explained by a linear model from extracted texture features.  
    To quantify the quality by which features encode each measure, we calculate the goodness of each fit using the coefficient of determination $R^2$.
    For \mbox{\clip} and \mbox{\clcs}, $r$ refers to the distance at which context sampling is performed, with nearest neighbor (NN) sampling being a special case of \mbox{\clcs}.
    The methods use different input modalities (Input) and produce features with different dimensionalities (Dim.).
    Bold values indicate the highest $R^2$ value per column.
  }
  \label{tb:regression}
  \begin{tabular}[]{lccc|cccc}
    \toprule
      Method & Dim. & $r$ [{\textmu}m] & Input & Cort. depth & WM depth & Curv. & Oblique. \\
    \midrule
      GLCM                                    & 36 & \multirow{4}{*}{-} & \multirow{4}{*}{$I_T$, $\sin\delta$, $\hat{\varphi}$} & 0.53 & 0.26 & 0.01 & 0.05 \\
      Histogram                               & 15 &  &  & 0.52 & 0.20 & 0.00  & 0.03 \\
      LBP                                     & 90 &  &  & 0.36 & 0.13 & 0.01  & 0.06 \\
      \mbox{[Histo., LBP, GLCM]}               & 141 &  &  & 0.63 & 0.27 & 0.02  & 0.08 \\
    \midrule
      \multirow{2}{*}{\mbox{\simclr}} & \multirow{2}{*}{2048} & \multirow{2}{*}{-} & $I_T$ & 0.60 & 0.22	& 0.00 & 0.18	\\
                                              &                       &                    & FOM   & 0.62 & 0.17 & 0.01 & 0.19 \\
    \midrule
      \multirow{3}{*}{\clip}  & \multirow{3}{*}{256} & 0   & \multirow{3}{*}{$I_T$, $\sin\delta$, $\varphi$} & 0.58 & 0.14 & 0.07  & 0.15 \\
                              &                      & 118 & & 0.79 & 0.34 & 0.13 & 0.43 \\
                              &                      & 236 & & 0.78 & \textbf{0.36} & 0.13 & 0.42 \\
    \midrule
    \multirow{3}{*}{\clcs}    & \multirow{3}{*}{256} & NN   & \multirow{3}{*}{$I_T$, $\sin\delta$, $\varphi$} & 0.79 & 0.35 & \textbf{0.17}  & 0.49 \\
                              &                      & 118 & & \textbf{0.82} & \textbf{0.36} & \textbf{0.17}  & \textbf{0.52} \\
                              &                      & 236 & & \textbf{0.82} & 0.34 & 0.14  & 0.51 \\
    \bottomrule
  \end{tabular}
\end{table*}

Fiber architecture has mutual dependencies with cortical morphology \cite{vanessen1997,striedter2015}.
To investigate to what extent different texture representations encode cortical morphology, we extract a range of morphological parameters from our test data. 
In particular, based on the cortex segmentation described in Sec.~\ref{sec:cortexsegmentation}, we compute a Laplacian field between outer pial and inner white matter surfaces using the \textit{HighRes cortex} \cite{leprince2015} module included in  \textit{brainvisa} \cite{cointepas2001}.
We extract the following measures:
\begin{itemize}[itemsep=1mm,leftmargin=.05\textwidth]
  \item \textit{Equivolumetric cortical depth} \cite{bok1929} as the depth along cortical traverses following the gradient of the Laplacian field, compensating for the effect of cortical curvature through the divergence of the same field. It has values of 0 at the Pial and 1 at the gray-white matter surface.
  \item \textit{White matter depth} defined as the shortest distance from each voxel within white matter to the interface between the cortical ribbon and white matter in millimeters.  
  \item \textit{Cortical curvature} as the divergence of the gradient of the Laplace field between Pial and white matter surfaces (\citet{goldman2005}, Equation 3.8; \citet{leprince2015})
  \item \textit{Obliqueness} of the sectioning plane, computed as the absolute angle between the gradient of the Laplacian field and the sectioning plane with values in [0°, 90°]. 
\end{itemize}

We follow the approach of \mbox{\citet{spitzer2020}} and quantify the extent to which these measures can be predicted from texture features by different methods using a linear model.
Being able to predict a quantity with a simple linear model indicates a robust encoding of that quantity.
We randomly select 10~000 voxels from the training set and compute both their feature representations using the trained models as well as the above-mentioned morphological measures.
The features are standardized using Z-score normalization and used to fit a linear regression model via least-squares.
For the model, we use relatively high L2 regularization with a weight of \mbox{10\textsuperscript{4}} to reduce overfitting observed for the pre-trained model (ImageNet) with high numbers of 2048 features.
The goodness of the fit is determined by calculating the coefficient of determination $R^2$, which denotes the proportion of variation in the measures that can be explained by the linear model from the feature representations.
We compute $R^2$ for predicted values from 10~000 randomly selected voxels from the test set.

Our results in \mbox{\cref{tb:regression}} show highest $R^2$ values for the prediction of all morphological measures from features by our \mbox{\clcs} method with medium sampling radius $r$ = 118.
It achieves $R^2$ values of 0.82 for cortical depth, 0.36 for white matter depth, 0.17 for curvature and 0.52 for obliqueness, which are overall higher compared to \mbox{\clip} with in-plane context sampling.
Using a larger or smaller radius for context sampling does not significantly increase $R^2$ values for \mbox{\clcs} or \mbox{\clip}.
While for \mbox{$r$ = 0}, i.e. using nearest neighbor (NN) context sampling for \mbox{\clcs}, $R^2$ values are marginally smaller, they decrease considerably for \mbox{\clip}, when using no context sampling in model training.
Classical texture descriptors (GLCM, Histogram, LBP) and the pre-trained model show overall much smaller $R^2$ values and do not encode curvature at all, i.e. showing values around 0.

\begin{figure*}[ht]
  \centering
  \begin{subfigure}[]{\textwidth}
    \centering
    \includegraphics[width=.99\textwidth]{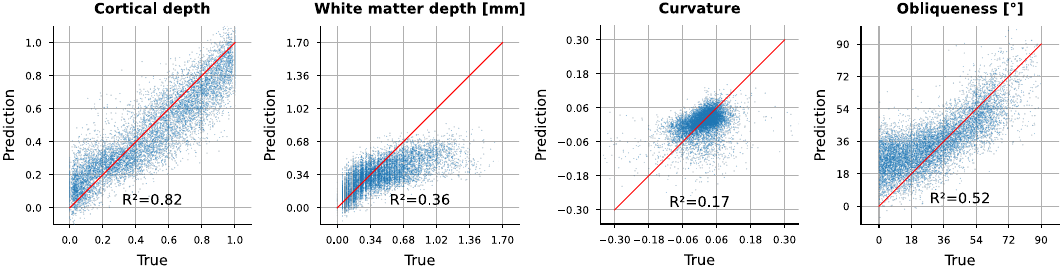}
    \caption{\clcs texture features}
    \label{fig:regression_ssl}
  \end{subfigure}
  \par\bigskip
  \begin{subfigure}[]{\textwidth}
    \centering
    \includegraphics[width=.99\textwidth]{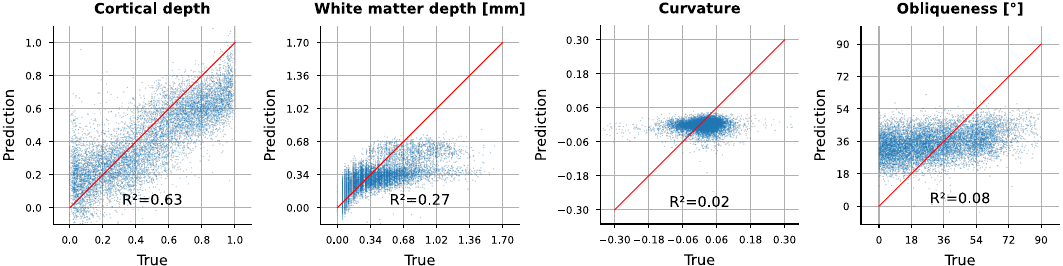}
    \caption{Combined classical texture features (Histogram, LBP, GLCM)}
    \label{fig:regression_glcm}
  \end{subfigure}
  \caption{
    Linear encoding of different morphological measures in (a) the proposed 256 dimensional \clcs feature representations and (b) 141 dimensional combination of classical texture features (Histogram, LBP, GLCM).
    A linear model is fitted via least squares to predict cortical depth, white matter depth, curvature, and obliqueness.
    Predicted and true target measures are shown as blue scatter plots, where red lines indicate an optimal fit.
    Goodness of each fit is calculated by the coefficient of determination $R^2$.
  }
  \label{fig:regression}
\end{figure*}

Scatter plots in \mbox{\cref{fig:regression}} visualize predicted and true values at the example of \mbox{\clcs} and combined classical texture features.
As shown in \mbox{\cref{fig:regression_ssl}}, for obliqueness, especially larger angles can be predicted from \mbox{\clcs} features, while the prediction of smaller angles in the scatter plot exhibits a deviation.
Predictions of white matter depth from features by both methods show a clearer fit for smaller depths than for larger depths.
As shown in \mbox{\cref{fig:regression_glcm}}, neither curvature nor obliqueness can be predicted from the selected classical texture features.

\subsection{Clustering of learned features}
\label{sec:clustering}

\subsubsection{Hierarchical clustering}
\label{sec:hierarch_clustering}
We perform hierarchical cluster analysis in the embedding space of \clcs features to evaluate the extent to which these features map characteristic nerve fiber configurations.
Hierarchical clustering creates a tree-like structure of hierarchical relationships among data points as a dendrogram based on their similarity in representational space.
We choose the bottom-up approach for agglomerative clustering, which merges closest clusters based on our choice of Euclidean distance and Ward linkage.
To reduce noise and computational effort, we cluster PCA-reduced feature maps from \cref{sec:pca}, where features are projected onto the first 20 components with 80.4\% total explained variance.
Additionally, using PCA-reduced feature maps mitigates potential negative effects of the curse of dimensionality when calculating distances in high-dimensional feature spaces.
To increase the receptive field and reduce in-plane noise, each feature map is smoothed by an in-plane 2D Gaussian kernel with a standard deviation of $\sigma=1$ which provides a good trade-off between noise reduction and keeping sensitivity to smaller structures.

\begin{figure}[p]
  \centering
  \includegraphics[width=.95\textwidth]{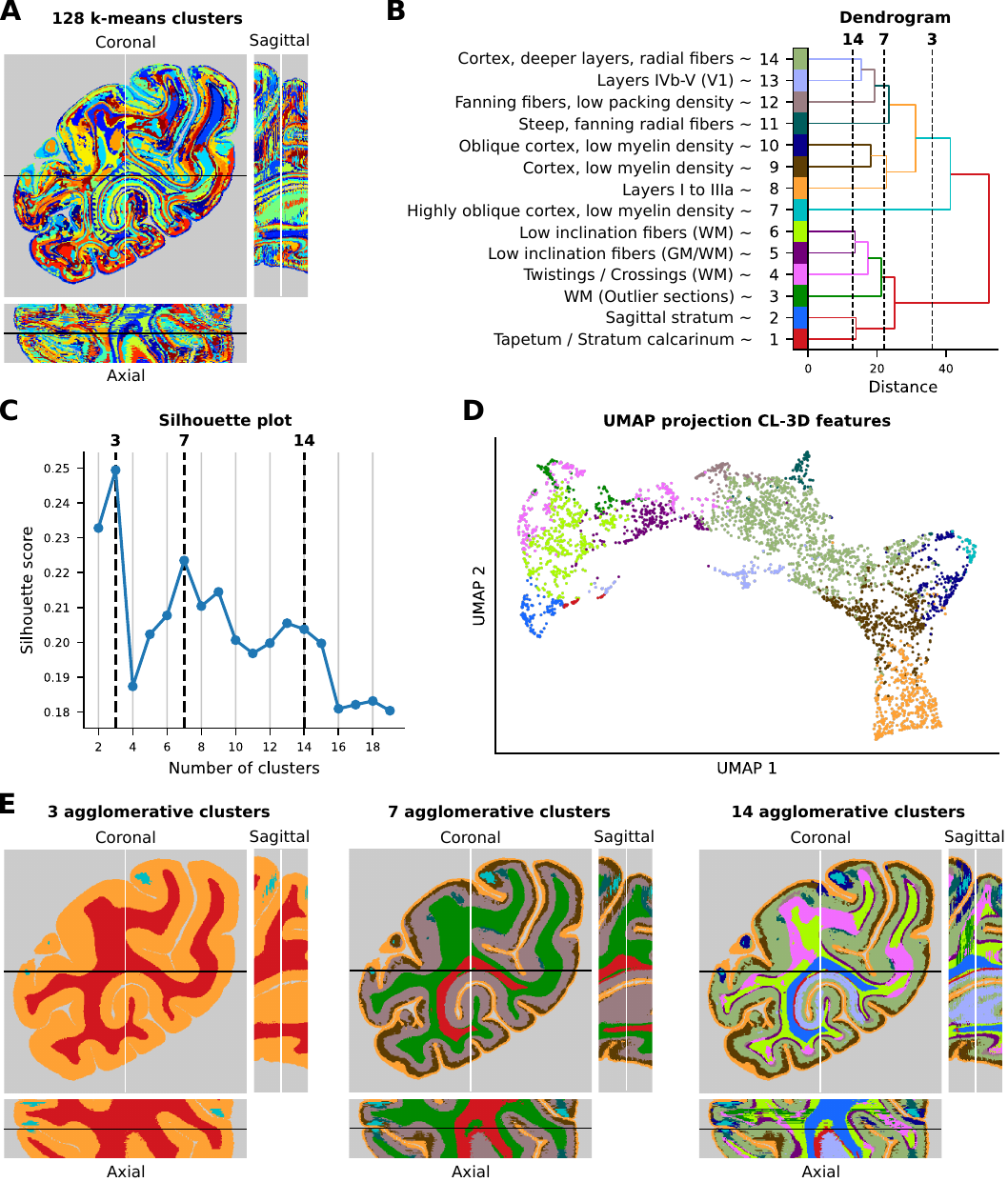}
  \caption{
    Agglomerative hierarchical clustering of 128 k-means centroids shows a hierarchy of fiber architecture.
    (A) 128 k-means centroids forming superpixel-like clusters.
    (B) Dendrogram representing distances between identified clusters before merging and approximate labels naming the structures with highest overlap with each cluster.
    (C) Silhouette plot showing local maxima around 3, 7 and 14 clusters.
    (D) UMAP projection of the \clcs features. The color of each point corresponds to the respective cluster assignment.
    (E) Clustering results for 3, 7 and 14 clusters.
  }
  \label{fig:clusters}
\end{figure}

We select features from foreground voxels across all test sections.
Due to the huge amount of 16 million data points, hierarchical clustering cannot be applied directly to the data points.
Instead, we perform a two-step approach by first performing k-means clustering on a subset of 100\,000 samples for 128 clusters.
We use the 128 resulting cluster centers to assign all remaining 15.9M data points to these clusters, allowing us to represent the whole test volume by superpixel-like clusters (\cref{fig:clusters}A).
Cluster assignments in the coronal plane appear visually smoother compared to the assignments across brain sections in the axial and sagittal planes.
As a second step, we perform agglomerative hierarchical clustering to obtain a cluster dendrogram (\cref{fig:clusters}B).
From this result, we calculate silhouette scores for an increasing number of clusters to identify interesting candidate clusterings for visualization (\mbox{\cref{fig:clusters}C}).
Considering the overall decline in silhouette scores, we observe local maxima at 3, 7, and around 13-14 clusters.
Notably, increasing from 13 to 14 clusters introduces a particularly interesting cluster, exclusively highlighting cortical layers in primary visual area (V1).
Therefore, we select 14 clusters for visualization.
We would like to point out that using fewer (32) or more (1024) k-means centroids for subsequent hierarchical clustering resulted in overall lower silhouette scores and led to unspecific, over-simplifying clusters or more noisy clusters, respectively.

Results for selected candidate cluster configurations for visualization are shown in \cref{fig:clusters}E.
We observe the same difference in spatial smoothness of cluster assignments between the coronal, axial, and sagittal planes as shown in \mbox{\cref{fig:clusters}A}, with the coronal plane appearing visually smoother.
A range of characteristic aspects of fiber architecture are revealed, which are identified and confirmed by two neuroanatomists (N.P.-G. and M.N.).
The descriptions are based on a comparison of each cluster with high-resolution \mbox{\mpli} images across multiple sections and its overall distribution within the 3D geometry of the brain to ensure the consistency of descriptions.

\textbf{3 clusters.}
Solutions for 3 clusters demonstrate a first global differentiation of the data into GM and WM.
Due to its high fiber density, cortical layer VI is sometimes represented inside the WM cluster.
We further observe a small cluster of tangentially cut cortex.

\textbf{7 clusters.}
The configuration with 7 clusters differentiates superficial and deep cortical layers.
This segregation is shaped by the packing density of radial fibers in the deeper layers, and the tangentially running fibers close to the pial surface.
For WM, voxels are split into two clusters:
1) the red cluster in \cref{fig:clusters}E highlights densely packed fibers with out-of-plane orientation of the sagittal stratum (SS), as well as surrounding densely packed in-plane fibers of the tapetum.
2) the green cluster in \cref{fig:clusters}E encompasses in-plane fibers or fibers with relatively low inclination, together with steep but less densely packed fiber bundles. 

\textbf{14 clusters.}
The configuration of 14 clusters reveals an increased sensitivity to specific WM fiber bundles and displays a cortical region delineation for the primary visual cortex (V1).
Furthermore, a range of fiber architectural properties can be recognized in the maps corresponding to the clusters in \cref{fig:clusters}B:
\begin{itemize}[itemsep=1mm,leftmargin=.05\textwidth]
  \item Cluster (1) shows the Tapetum and stratum calcarinum (SC), characterized by approximate in-plane fibers with high packing density.
  \item Cluster (2) displays densely packed, highly inclined fibers and fibers of the sagittal stratum (SS) with out-of-plane orientation.
  \item Cluster (3) includes WM voxels from sections with an artifact-related increased light transmittance (not present in the section shown in \cref{fig:clusters}E).
  \item Cluster (4) mainly displays WM voxels, but in some cortical segments also encompasses layer VI. WM covered by these voxels is characterized by fibers with high packing density, small inclination angles, and twisting or crossing patterns. 
  \item Cluster (5) mainly highlights layer VI of cortical segments complementing cluster 4. WM voxels encompass fibers with low packing density and low inclination at the border between GM/WM.
  \item Cluster (6) reveals only WM voxels with very small fiber inclination mostly parallel configurations, with only a few crossings.
  \item Cluster (7) highlights voxels located in highly oblique cortex with layers characterized by low myelination. When not tangentially sectioned, this portion of the cortex is encompassed by cluster (9).
  \item Cluster (8) highlights the most superficially located bands of tangential fibers, namely those of the zonal layer and the Kaes–Bechterew stripe, which are located in cytoarchitectonic layers I and IIIa, respectively \cite{zilles2015}, and found throughout the whole cortex.
  \item Cluster (9) covers cortical layers with low density of myelin. The width of this cluster varies along the cortical ribbon. Some areas, where radial fibers reach almost up to layer II, are characterized by a narrow band of cluster (9), while in other areas it is broad because their radial fibers only reach into layer IIIb. The cluster disappears in the cortex of highly compressed sulci.
  \item Voxels in cluster (10) are also located in obliquely sectioned layers with a low myelination while being less oblique than those of cluster (7).
  \item Cluster (11) encompasses steep radial fibers fanning out at the apex of the gyrus.
  \item Cluster (12) highlights approximate in-plane fibers with low packing density, fanning out at the apex of the gyrus.
  \item Voxels of cluster (13) are restricted to the primary visual area (V1). They are mainly found in layers IVb-V, but, depending on packing density, sometimes also those of layer VI). This cluster could reflect local cortical connectivity within V1.
  \item Cluster (14) highlights deeper layers mainly due to their density of radial fibers. Its width varies along the cortical ribbon and is inversely related to the width of cluster (9). In V1 cluster (14) is restricted to layers IIIb and IVa. In the remaining cortex, depending on the area it can reach from layer IIIa or IIIb to layer V or VI.
\end{itemize}

To visualize the organization of data points in the learned representational space, we additionally perform UMAP \cite{mcinnes2018} projection of the PCA-reduced features used for clustering on two dimensions for 4\,000 example data points.
Projections in \cref{fig:clusters}D show the organization of features in a continuous band along cortical depth starting from superficial layers (clusters 8 and 9), to deeper layers (clusters 13, 14, 5, and partially 4) until reaching WM clusters (3, 6 and partially 4).
Branches to the sides highlight different degrees of obliqueness of cortical layers (i.e. clusters 7, 9, and 10).
Clusters for structures such as the SS (cluster 2), Tapetum/SC (cluster 1), or layers of V1 (cluster 13) form separate branches in the projected space.
We observe splits into multiple fragments by clusters 3, 4 and 13 in UMAP space.
These splits can be partially explained by the features providing a more fine-grained distinction between patches from different cortical and white matter depths than captured by the clustering results.
While increasing the number of clusters in hierarchical clustering reveals these additional details of fiber architecture, we limit the number of clusters to 14 in our qualitative analysis to maintain interpretability.
Beyond this point, the complexity of clusterings gradually increases, and the content of individual clusters becomes more difficult to describe.

\subsubsection{Consistency of cluster assignments across sections}

\begin{figure*}[t]
  \centering
  \includegraphics[width=.95\textwidth]{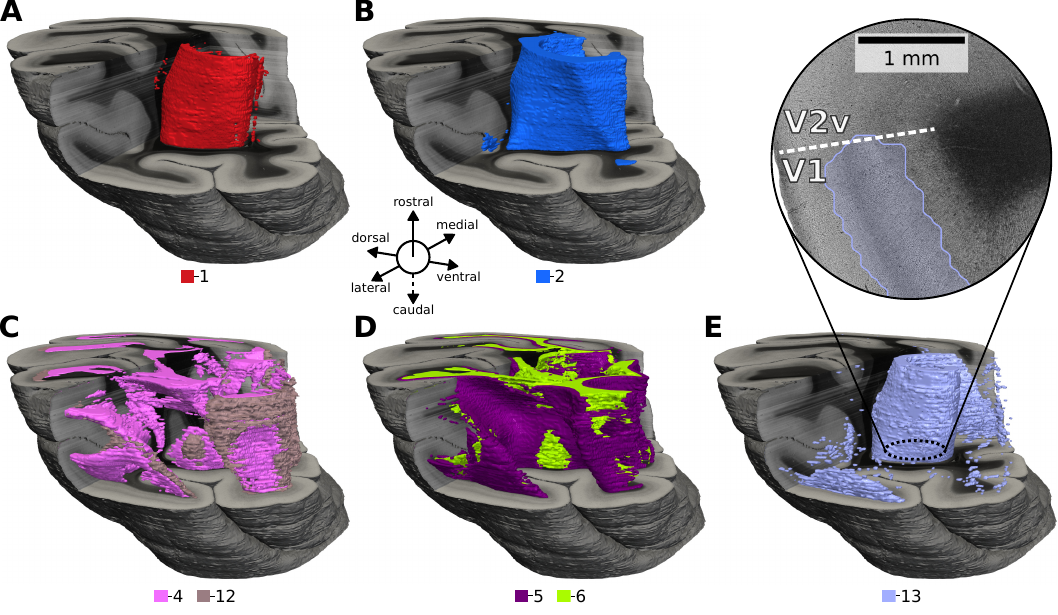}
  \caption{
    Hierarchical agglomerative clustering of representations extracted by the \clcs approach produces consistent 3D segments with anatomical relevance.
    (A) Approximately the Tapetum and stratum calcarinum (1).
    (B) Contains steep fibers of the sagittal stratum (2).
    (C) Close to in-plane twisting or crossing fibers (4) and flat fanning fibers (12).
    (D) Flat fibers for layer VI (5) and WM (6).
    (E) Mainly layers IVb-V of primary visual area (V1), sometimes including layer VI (13).
    It is less clearly defined than other clusters.
  }
  \label{fig:clusters_3d}
\end{figure*}

In addition to 2D cluster maps, we assemble 3D renderings of the configuration with 14 clusters (\cref{fig:clusters_3d}).
While \clcs features were generated based only on in-plane texture information without cross-section constraints, the volume renderings reveal consistent cluster boundaries across sections, as can be observed by the smooth cluster shapes in the cross-sections.
Cluster 13 stands out in being more noisy than other clusters.

To better quantify the cross-section consistency of feature clusters, we compute the intersection over union (IoU) of cluster assignments between adjacent sections for different numbers of k-means clusters (2, 8, 32 and 128).
Since absolute coordinates are not included in the features, this analysis provides insights into the robustness of representations to inter-section variations arising from histological processing.

\begin{table*}[t]
  \centering
  \caption{
    Cross-section consistency as mean IoU of k-means cluster assignments between neighboring sections based on different texture features.
    For \mbox{\clip} and \mbox{\clcs}, $r$ refers to the distance at which context sampling is performed, with nearest neighbor (NN) sampling being a special case of \mbox{\clcs}.
    Column "Input" denotes modalities used as input for each method.
    Using a cross-section sampling strategy in \clcs achieves the overall highest consistency.
    Bold values indicate the highest mean IoU score per column.
  }
  \label{tb:cross_sections}
  \begin{tabular}[t]{lcc|cccc}
    \toprule
      Method & $r$ [\mmu{}] & Input & 2 clusters & 8 clusters & 32 clusters & 128 clusters \\
    \midrule
      GLCM & \multirow{4}{*}{-} & \multirow{4}{*}{$I_T$, $\sin\delta$, $\hat{\varphi}$} &  95.2 & 58.4 & 34.1 & 19.4 \\
      Histogram                   & & & 95.4 & 49.8 & 27.7 & 14.3 \\
      LBP                         & & & 49.4	& 30.9 & 17.3 & 9.2	 \\
      \mbox{[Histo., LBP, GLCM]}   & & & 92.9 & 42.6 & 24.2	& 14.1 \\
    \midrule
      \multirow{2}{*}{\mbox{\simclr}} & \multirow{2}{*}{-} & $I_T$  & 86.6 & 50.0	& 29.9 & 17.0	\\
      &  & FOM & 77.8 & 50.9 & 35.5	& 23.6 \\
      
    \midrule
      \multirow{3}{*}{\clip} & 0   & \multirow{3}{*}{$I_T$, $\sin\delta$, $\varphi$} &88.9 & 47.1 & 25.0 & 12.8 \\
                             & 118 & & 95.4 & 61.2 & 35.9 & 20.0 \\
                             & 236 &  & 95.0 & 55.8 & 36.7 & 21.0 \\
    \midrule
      \multirow{3}{*}{\clcs} & NN & \multirow{3}{*}{$I_T$, $\sin\delta$, $\varphi$} & 89.2 & 70.8 & 50.2 & 30.3 \\
                             & 118 & & \textbf{96.1} & 70.5 & \textbf{51.0} & \textbf{32.3} \\
                             & 236 & &95.4 & \textbf{71.9} & 50.7 & 32.0 \\
    \bottomrule
  \end{tabular}
\end{table*}

IoU scores reported in \mbox{\cref{tb:cross_sections}} decrease overall as the number of clusters increases.
Cluster assignments based on \mbox{\clcs} features are significantly more consistent across all numbers of clusters compared to the other methods.
When using nearest neighbor (NN) context sampling for \mbox{\clcs}, scores slightly decrease.
IoU scores of other methods (\mbox{\clip}, GLCM, pre-trained encoder on \mbox{ImageNet}) are relatively close to each other.
Setting \mbox{$r$ = 0} for \mbox{\clip}, i.e. not performing context sampling, stands out due to particularly low IoU values, which fall below those of GLCM and the pre-trained encoder on \mbox{ImageNet} for more than 2 clusters.

\subsection{Using \clcs features for retrieval of common fiber orientation patterns}

\begin{figure}[p]
  \centering
  \includegraphics[width=.95\textwidth]{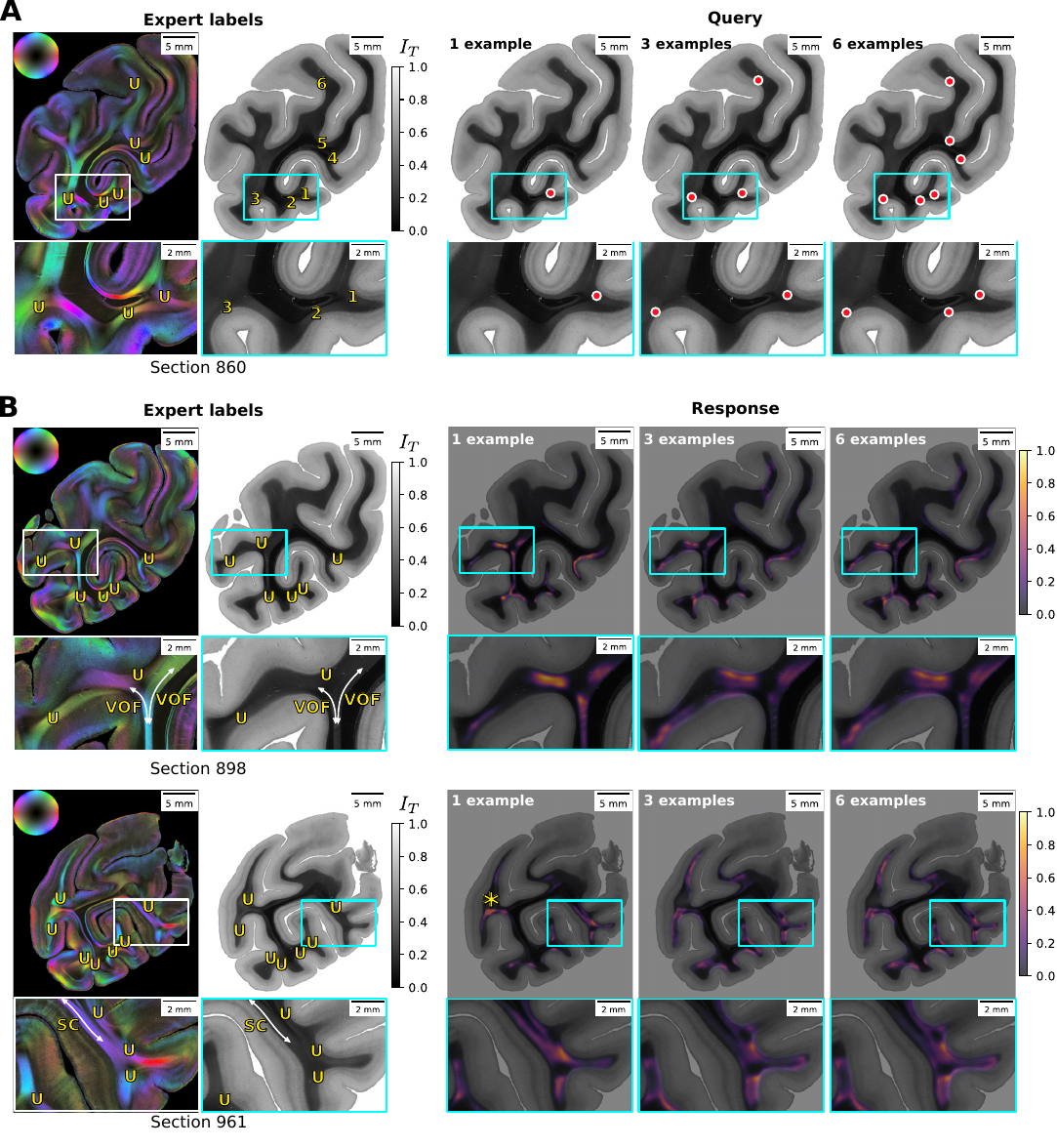}
  \caption{
    \clcs features allow retrieval of distinct nerve fiber patterns by using prototypical patches as queries and searching for nearby patches in feature space.
    (A) A section from the test set with query points that identify U-fibers.
    Yellow 'U' labels in fiber orientation maps (left) and numbers in transmittance images (right) show identified U-fibers.
    Red dots mark query points for U-fibers with increasing examples (1, 3, and 6).
    (B) shows responses to queries in two other sections from the test set, displaying similarities of voxels to the mean representation of example points as affinity maps.
    The asterisk highlights the position of a U-fiber not detected by this procedure.
    U: U-fibers,
    VOF: vertical occipital fascicle,
    SC: stratum calcarinum.
  }
  \label{fig:distance_function}
\end{figure}

The proposed \clcs method embeds image patches with similar fiber architectural properties as close points in the learned feature space.
Therefore, we expect that the resulting feature representations can be used for retrieval of similar structures, given a known "prototype" image patch. 
To evaluate the suitability for such an application, we investigate how far U-fiber structures can be found in \mpli image data from a few image examples.
Although U-fibers are represented by one of the principal axes (\cref{fig:pca}D (7)), they do not appear as individual clusters in \cref{fig:clusters}E.
To demonstrate that the representations can still be used to identify specific nerve fiber architectures, we provide a few positive examples of U-fibers to perform a query for similar fiber configurations (\cref{fig:distance_function}).

Unfortunately, access to detailed, annotated data in combination with \mpli measurements is limited, as reliable identification of positive and negative examples is not always possible.
However, we can still search for U-fibers by selecting only a few positive examples that could be identified with confidence from \mpli images by \citet{takemura2020}.
For this purpose, we select up to 6 query points as positive examples (\cref{fig:distance_function}A) and compute affinity maps that show the similarity of all voxels in the feature maps to the averaged query features in representational space as responses.
For this experiment, we smooth PCA-reduced feature maps with 20 components (80.4\% total explained variance) by a 2D Gaussian kernel analog to \cref{sec:clustering}, but set $\sigma$ = 2 to increase the receptive field of texture features.
This allows to represent texture for larger structures such as U-fibers and improves retrieval results.
We calculate the affinity between feature points using a Gaussian radial basis function (RBF) kernel with $\sigma$ = 3.5.

The affinity maps reveal peak regions at locations displaying U-fiber structures (\cref{fig:distance_function}B).
They activate to all U-fibers identified by an expert, except for one missing activation for a U-fiber, highlighted by the asterisk.
In addition, the activations highlight some fiber bundles that are not labeled as U-fibers, such as the vertical occipital fascicle (VOF) or the stratum calcarinum (SC).
Although the false positive activations for VOF and SC could be suppressed through more query points, they do not disappear completely.

  \section{Discussion}
\label{sec:discussions}
\subsection{\clcs features encode fundamental aspects of fiber architecture}

Feature representations extracted by the proposed \clcs method from \mpli image patches encode distinct aspects of fiber architecture.
Our experiments in Sec.~\ref{sec:clustering} showed that the features form hierarchical clusters that represent gray and white matter, myeloarchitectonic layer structures, fiber bundles, fiber crossings, and fiber fannings.
These clusters are spatially consistent and often highlight specific characteristics of fiber architecture as locally connected structures (\cref{fig:clusters}).
Even without explicit clustering, the main PCA components of the feature embedding space produce maps that highlight fundamental principles of fiber architecture (\cref{fig:pca}).
This indicates that proximity of features in the embedding space, which is efficient and easy to compute, serves as a suitable proxy measure for similarity of fiber architecture as captured in the corresponding image patches. 
This is in contrast to directly measuring similarity of \mpli image patches, where strong differences can occur despite very similar fiber configurations.
Thus, \clcs features are well suited to facilitate downstream applications for \mpli image analysis. 

\subsection{\clcs features are robust to variations in histological processing}

While encoding relevant aspects of fiber architecture, the proposed \clcs features proved to be robust against many other sources of texture variation.
We were able to observe this robustness in the 3D stacking of consecutive images with derived cluster segments from \clcs features. 
The segments showed a high overlap across brain sections (\mbox{\cref{tb:cross_sections}}), in particular, compared to the clustering of classical texture features and a pre-trained encoder on ImageNet as baselines, but also w.r.t.~\mbox{\clip} features.
Clustering results by these methods were overall not consistent enough (see \mbox{\cref{fig:clustering_baslines_appendix}}) to perform the same in-depth qualitative evaluation as performed for \mbox{\clcs} in \mbox{\cref{sec:hierarch_clustering}}.
Volume renderings in \cref{fig:clusters_3d} illustrated the consistency of clustering \clcs features as spatially consistent 3D segments.
This suggests that \clcs promotes the learning of representations that are more robust to discrepancies between independently processed sections.
\clcs features were also found to be robust to the absolute in-plane orientation of texture, as shown by consistent laminar patterns of cluster assignments regardless of their absolute orientation in 2D (\cref{fig:clusters}).

Some of the robustness of \mbox{\clcs} features can be attributed to the introduced context sampling across brain sections, as shown in \mbox{\cref{tb:cross_sections}}.
\mbox{\clcs} features demonstrated significantly higher overlap in cluster assignments between sections compared to models trained with in-plane (\mbox{\clip}) or without context sampling (\mbox{\clip} with \mbox{$r$ = 0}).

Another factor contributing to the robustness may result from the data augmentations specifically designed for \mbox{\mpli} parameter maps, introduced in \mbox{\cref{sec:transformations}}.
\mbox{\clcs} and \mbox{\clip} models trained with the introduced augmentations showed increased feature quality and robustness in \mbox{\cref{sec:lin_ev_ablation}}.
A comparison of individual augmentations identified the highest benefit from using color distortions (modulation of section thickness and the attenuation coefficient).
For \mbox{\clip}, using the blur augmentation alone performed worse than training a model without augmentations.
This is in line with \mbox{\citet{chen2020a}}, who found the highest benefit from color distortions and the least benefit from blur.
For \mbox{\clcs}, geometric affine and flip augmentations each demonstrated a negative effect when used individually as the only augmentation.
This is surprising at first glance but could be explained by natural geometric distortions in the training pairs, which were sampled from unregistered, neighboring tissue sections.
Including geometric transformations to the full set of augmentations, on the other hand, did not negatively impact \mbox{\clcs} training.

Remaining inconsistencies in cluster assignments between sections as shown in the sagittal and axial planes in \mbox{\cref{fig:clusters}A} and \mbox{\cref{fig:clusters}E}, can be attributed to some variations between brain sections still captured in \mbox{\clcs} features.
For example, the "WM (Outlier sections)" cluster highlights white matter in sections with degraded transmittance, which we regard as a histological artifact.
Imprecisions in the 3D reconstruction of adjacent brain sections could also contribute to this effect.
The cross-section discontinuity of cluster assignments could be mitigated by post-processing with the same spatial smoothing of features in the axial and sagittal planes as performed for feature maps in the coronal plane.

\clcs and \clip features showed sensitivity to the relative cutting angle of cortical voxels, as observed by their ability to predict measured obliqueness using a linear model (\cref{tb:regression}).
Obliqueness is a local feature of histological images that is consistent across adjacent sections and could be exploited by the contrastive learning objective to identify nearby positive pairs.
For the majority of cortical voxels, however, this effect on the features seems to be small.
Only for very high obliqueness \clcs features formed some smaller branches in the UMAP projection in \cref{fig:clusters}D or isolated clusters in \cref{fig:clusters}E.
This aligns with observations from the scatter plots (\cref{fig:regression_ssl}) indicating that \clcs features primarily encode obliqueness for larger cutting angles, showing limited ability to predict smaller angles.
If an encoding of obliqueness in downstream applications is nevertheless not intended, a supervisory signal that combines texture patches with different cutting angles into the same label could be helpful.
For unsupervised learning, treating obliqueness as a confounding variable \cite{snoek2019,dinga2020} could also help to reduce its effect.

\subsection{Retrieval and mapping as possible downstream applications}

Fiber architecture is expressed in highly complex textures when measured at microscopic resolution.
This makes it extremely challenging to navigate and explore larger stacks of \mpli images. 
An obvious, albeit simple, application is the search for similar local configurations of nerve fibers, given a template image patch used as a prototype for such a query.
We took a search for U-fiber structures as an example (\cref{fig:distance_function}), where independent expert annotations could be obtained from a previous study, and were able to demonstrate the feasibility of such a retrieval task with the proposed \clcs features.

We showed that features cluster into groups that follow certain fiber bundles such as the sagittal stratum or the Tapetum (\cref{fig:clusters}).
While this might at first suggest the use of features for automated brain mapping tasks, the clusters did not lend themselves to a sufficiently accurate delineation of anatomical structures.
This could be due to partial volume effects of patches used to represent texture, or to the smoothing performed before clustering to denoise the features.
As contrastive learning focuses on the most characteristic properties of texture to identify positive pairs, some aspects of fiber architecture might overshadow others.
The clustering in \cref{fig:clusters}E, for example, did not fall into accurate GM/WM segments.
This could be due to features reflecting the density of myelinated fibers more than other aspects of fiber architecture, thus including the deepest cortical layers in the WM segment, where the density of myelinated fibers is still very high.
Since the degree of sharpness of the boundary between cortex and white matter constitutes a criterium for the identification of architectonically distinct areas \cite{niu2020}, another explanation would therefore be that brain areas with a blurry boundary are those in which voxels of layer VI merge into the WM segment.
It should be noted though, that the proposed feature extraction method is not specifically designed for performing automatic brain mapping.
A supervised approach for brain mapping as a downstream application can nevertheless be promising with the proposed features.
Linear evaluation, as shown in \mbox{\cref{fig:lin_ev_methods}}, demonstrated that a linear classifier on top of \mbox{\clcs} or \mbox{\clip} features required only 30 examples per tissue class (gray matter, white matter or background) to perform convincing classification, significantly reducing the amount of manual annotation required.
A more systematic investigation into fiber architectonic mapping based on \clcs features will be an important follow-up work of this study.

\subsection{Relationship of feature representations with brain morphology}

Cortical layers are arranged along the cortical depth and have distinct characteristic architectures.
Being able to regress cortical depth from texture features using a simple linear model therefore suggests that features robustly encode information about the layering structure of the cortex, which is important for downstream applications such as brain mapping.
The high amount of variance in cortical depth that could be explained by a linear model from \clcs and \clip features (\cref{tb:regression}) indicates that these models indeed encode layer-related textures.
For \clcs, this claim is also supported by the observation that the main PCA components highlight individual cortical layers (\mbox{\cref{fig:pca})}, while clustering of the features shows several clusters that group more superficial and deeper layers \mbox{(\cref{fig:clusters})}.
Furthermore, being able to regress the depth of patches within white matter (\cref{tb:regression}) suggests that \clcs and \clip features robustly separate deep from superficial fiber bundles such as U-fibers \cite{decramer2018,shinohara2020}.
Classical texture features, as well as a pre-trained model on ImageNet, on the other hand, demonstrated significantly lower capability in predicting cortical and white matter depth.
A \mbox{\clip} model with sampling radius $r$ = 0, i.e. not performing context sampling, performed much worse.
This highlights a positive effect of the introduced context sampling in learning expressive representations for fiber architecture in \mbox{\mpli}.

Methods for analyzing the laminar structure of the cortex typically require representations that are robust to cortical folding \cite{schleicher1999,waehnert2014,leprince2015}.
While \clcs and \clip learned to represent some curvature-related patterns such as for fanning radial fibers at gyral crowns (\cref{fig:clusters}), predicting curvature from the features did not work well (\cref{tb:regression}), indicating moderate robustness of the proposed feature representations to cortical folding.
For \clcs and \clip, the weak encoding might be attributed to the affine transformation applied in the contrastive learning setup, which performs scaling and shearing operations on the texture examples.
It should be noted that in addition to the curvature definition used in \cref{sec:linear_relationships}, other established definitions \cite{goldman2005} that have not been considered in this study might lead to different results.

  \section{Conclusion and Outlook}
\label{sec:conclusion}

Aiming to improve automatic mapping and analysis of fiber orientation distributions in the brain, we introduced a self-supervised contrastive learning scheme for extracting "deep" feature representations for \mpli image patches at micrometer resolution.
We specifically proposed \textit{3D context Contrastive Learning} (\clcs), introducing a \textit{context sampling} strategy to sample positive pairs based on their spatial proximity across nearby brain sections.
Without any anatomical prior information given during training, the feature representations extracted by \clcs were shown to highlight fundamental patterns of fiber architecture in both gray and white matter, such as myelinated radial and tangential fibers within the cortex, fiber bundles, crossings, and fannings.
At the same time, feature representations by \clcs proved to be more robust to variations between independently measured sections, such as artifacts arising from histological processing, compared to statistical methods, an encoder pre-trained on natural images, and representations by in-plane sampling of positive pairs in contrastive learning (\clip).

The present study opens new perspectives for automated analysis of fiber architecture in \mpli.
Due to the low-dimensional embedding space, \clcs feature representations can aid in interpretation of \mpli textures and improve computational efficiency of downstream \mpli analysis.
For example, the learned feature representations can be used to develop spatial maps of specific aspects of fiber orientation distributions, such as U-fibers, which allow comparison of fiber architecture with other modalities linked to brain atlases.
They can also be used to train discriminative models for downstream tasks such as segmenting tissue classes, cortical layers, fiber bundles, or even brain areas with minimal amount of positive and negative labelled examples.

An important direction for future research will be to extend the trained models to larger training datasets.
We intend to extend the approach to whole brain datasets, possibly including multiple species.
Since the main challenge for establishing training data is the precise 3D reconstruction from individual brain sections, it will be helpful to investigate how far approximate registrations can be sufficient for \clcs.
Furthermore, we plan to integrate deep \mpli features into brain atlases to provide easy accessibility.
For the human brain, the BigBrain \cite{amunts2013} model would be an ideal reference model for integration, which is already used for multimodal data integration from other imaging modalities, such as cyto- and receptor architecture.

  \section*{Ethics Statement}
\label{sec:ethics}
Vervet monkeys used were part of the Vervet Research Colony housed at the Wake Forest School of Medicine.
Our study did not include experimental procedures with living animals.
Brains were obtained when animals were sacrificed to reduce the size of the colony, where they were maintained and sacrificed in accordance with the guidelines of the Wake Forest Institutional Animal Care and Use Committee IACUC \#A11-219 and the AVMA Guidelines for the Euthanasia of Animals.
  
  \section*{Data and Code Availability}

The software used to implement and train self-­supervised \clcs and \clip models is available on GitHub\footnote{\url{https://github.com/FZJ-INM1-BDA/cl-3d}}.
An implementation of the introduced data augmentations for \mpli images\footnote{\url{https://jugit.fz-juelich.de/inm-1/bda/software/data_processing/pli-transforms}} as well as additional dependencies \footnote{\url{https://jugit.fz-juelich.de/inm-1/bda/software}} are available on GitLab.

Volumetric clustering results, PCA projections of extracted \clcs features, measures of cortex morphology, and \mpli fiber orientation and transmittance maps for reference are available on the EBRAINS data sharing platform \mbox{\cite{oberstrass2024c}}.

The complete set of high-resolution \mpli images for the vervet monkey occipital lobe is hosted on the Jülich supercomputing facility.
A subset of selected \mpli images is available on EBRAINS \cite{axer2020a}, with a future publication of the whole stack of images planned.

\section*{Author Contributions}

\textbf{Alexander Oberstrass:}
Conceptualization,
Methodology,
Software,
Formal analysis,
Investigation,
Data curation,
Writing - original draft,
Visualization.
\textbf{Sascha E. A. Muenzing:}
Methodology,
Data curation,
Writing - review \& editing.
\textbf{Meiqi Niu:}
Validation,
Data curation,
Investigation,
Writing - review \& editing.
\textbf{Nicola Palomero-Gallagher:}
Validation,
Investigation,
Writing - original draft,
Writing - review \& editing.
\textbf{Christian Schiffer:}
Conceptualization,
Methodology,
Software,
Writing - review \& editing.
\textbf{Markus Axer:}
Conceptualization,
Validation,
Investigation,
Supervision,
Writing - review \& editing,
Project administration,
Funding acquisition.
\textbf{Katrin Amunts:}
Conceptualization,
Validation,
Investigation,
Supervision,
Writing - Review \& Editing,
Resources,
Project administration,
Funding acquisition.
\textbf{Timo Dickscheid:}
Conceptualization,
Methodology,
Supervision,
Writing - original draft,
Writing - Review \& Editing,
Project administration,
Funding acquisition.

\section*{Funding}
\label{sec:funding}
This project received funding from the Helmholtz Association’s Initiative and Networking Fund through the Helmholtz International BigBrain Analytics and Learning Laboratory (HIBALL) under the Helmholtz International Lab grant agreement InterLabs-0015, the Helmholtz Association portfolio theme ``Supercomputing and Modeling for the Human Brain'', and the European Union’s Horizon 2020 Research and Innovation Programme, grant agreement 945539 (HBP SGA3), which is now continued in the European Union’s Horizon Europe Programme, grant agreement 101147319 (EBRAINS 2.0 Project). 
Computing time was granted through JARA on the supercomputer JURECA at Jülich Supercomputing Centre (JSC).
Vervet monkey research was supported by the National Institutes of Health under grant agreements R01MH092311 and P40OD010965.

\section*{Declaration of Competing Interests}
\label{sec:conflict_of_interest}
The authors declare that the research was conducted in the absence of any commercial or financial relationships that could be construed as a potential conflict of interest.

\section*{Acknowledgements}
\label{sec:acknowledgements}
We sincerely thank Karl Zilles and Roger Woods for their valuable collaboration in the vervet brain project,
the lab team of the Institute of Neuroscience and Medicine (INM-1) for preparing and measuring the brain sections,
and the members of the Big Data Analytics and Fiber Architecture groups (both INM-1) for their valuable inputs and discussions during the development of this research.

\ifx\documenttype\arxiv
  \bibliographystyle{apalike}
  \bibliography{bib/refs_zotero.bib,bib/refs_manual.bib}
\fi

\clearpage

\appendix

\section{Appendix}

\begin{figure}[h]
  \centering 
  \includegraphics[width=.55\textwidth]{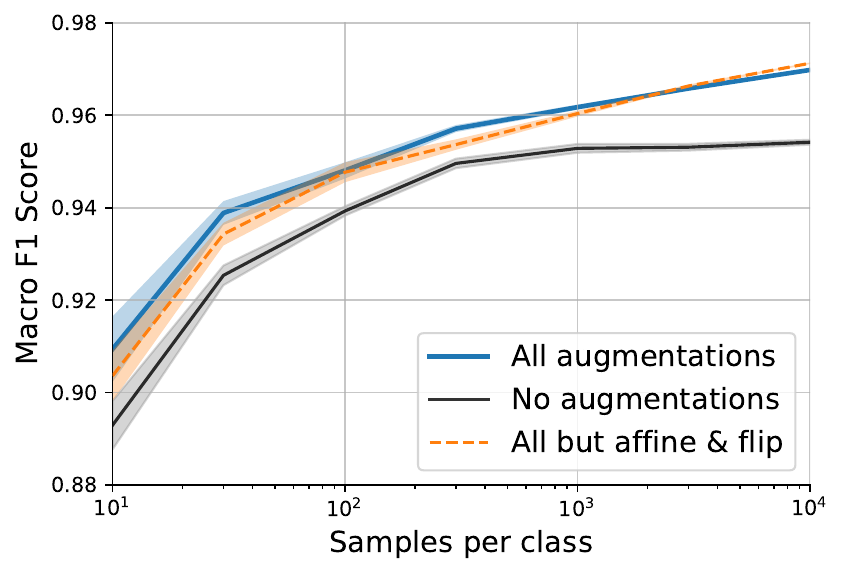}
  \caption{
    Excluding geometric affine and flip data augmentations from \mbox{\clcs} training does not improve model performance.
    Quality and robustness of features are evaluated using the linear evaluation protocol under an increasing amount of labeled samples per class.
    A simple linear classifier is fitted on extracted features to differentiate texture patches as gray matter, white matter or background.
    Macro F1 scores are presented for each model across different augmentation sets.
    Shaded areas indicate the standard error over 50 independent runs with random training samples.
  }
  \label{fig:lin_ev_geom_cl3d}
\end{figure}

\begin{figure*}[p]
    \centering
    \includegraphics[width=.95\textwidth]{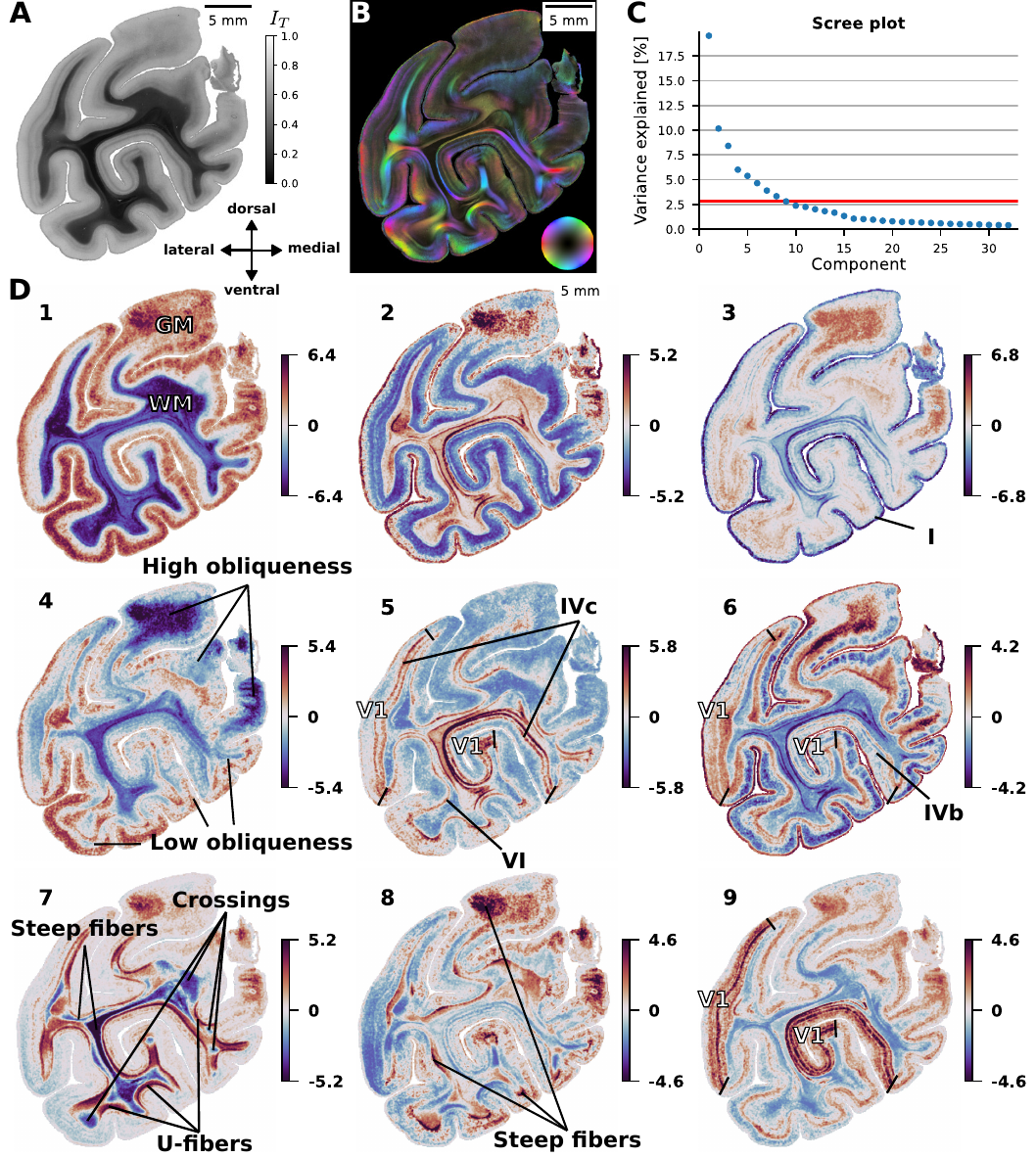}
    \caption{
      Projection of \clcs features from section 961 onto the 9 PCA components with largest explained variance.
      (A) Transmittance and (B) fiber orientation maps (FOM).
      (C) Scree plot showing the variance explained by the first 32 components. The horizontal red line indicates the variance explained by the 9th component of 2.8\%.
      (D) Color-coded parameter maps of the selected PCA components, with background pixels masked as zero. The maps reveal anatomically plausible structures.
      Structures identified by each PCA component in this section are consistent with section 898 (\mbox{\cref{fig:pca}}).
      GM: gray matter, WM: white matter, V1: primary visual cortex.
    }
    \label{fig:pca_appendix}
  \end{figure*}

\begin{figure}[p]
    \centering 
    \includegraphics[width=.95\textwidth]{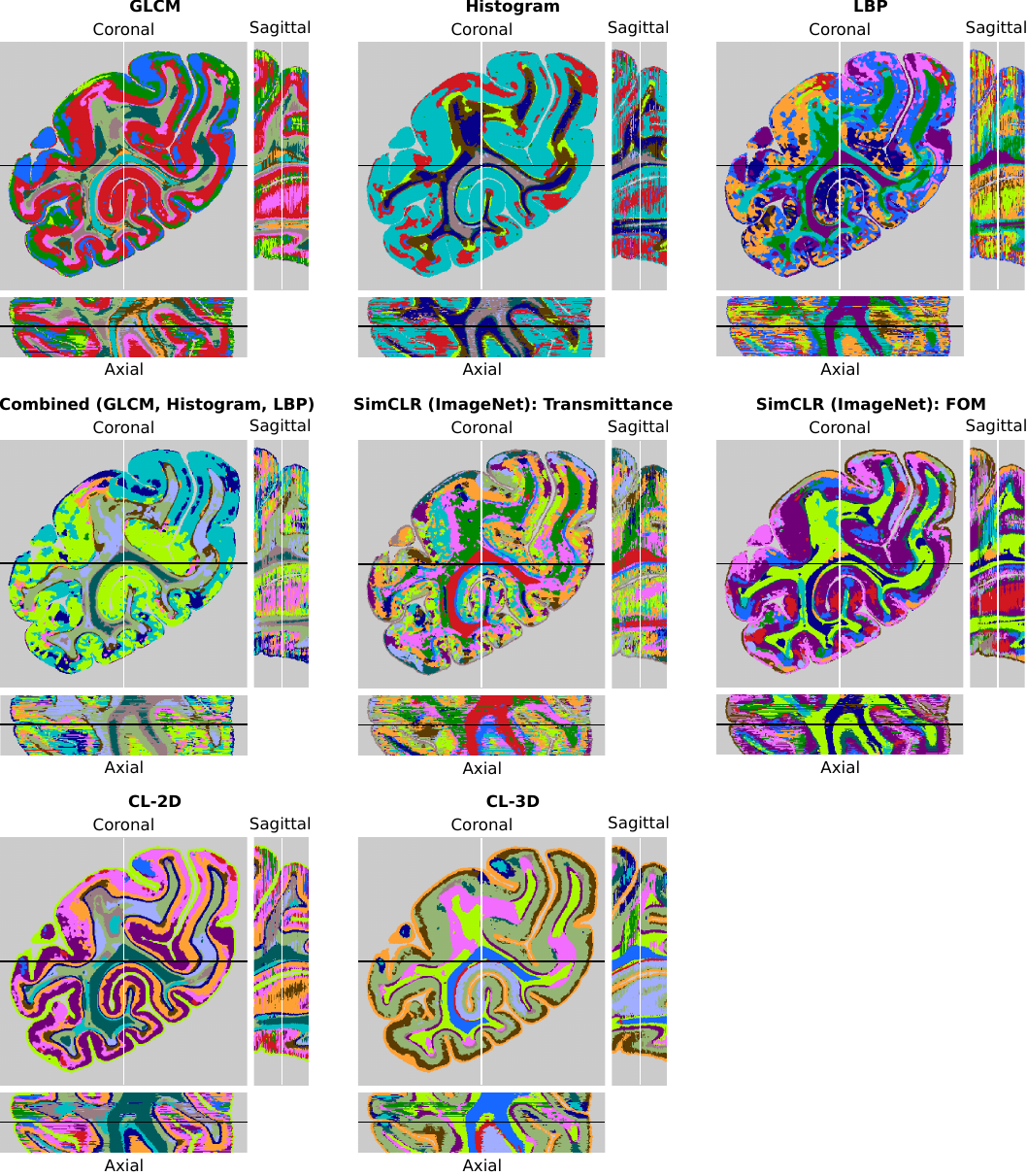}
    \caption{
      Agglomerative hierarchical clustering results for 14 clusters across all feature extraction methods show highest cluster quality by \mbox{\clcs}.
      Baseline GLCM, Histogram, LBP and combined features, as well as a pre-trained encoder on ImageNet using transmittance and FOM images, produce fragmented cluster assignments in the coronal plane and inconsistent assignments in the sagittal and axial planes.
      In contrast, \mbox{\clip} and \mbox{\clcs} demonstrate more organized cluster assignments in the coronal plane, with \mbox{\clcs} showing most consistent cluster assignments in the axial and sagittal planes.
    }
    \label{fig:clustering_baslines_appendix}
  \end{figure}

\end{document}